\documentclass{article}

\PassOptionsToPackage{numbers, sort, compress}{natbib}
\usepackage[final]{neurips_2021}

\usepackage[utf8]{inputenc} % allow utf-8 input
\usepackage[T1]{fontenc}    % use 8-bit T1 fonts
\usepackage[pagebackref=true,breaklinks=true,letterpaper=true,colorlinks,bookmarks=false]{hyperref}
\usepackage{url}            % simple URL typesetting
\usepackage{booktabs}       % professional-quality tables
\usepackage{amsfonts}       % blackboard math symbols
\usepackage{nicefrac}       % compact symbols for 1/2, etc.
\usepackage{microtype}      % microtypography
\usepackage{xcolor}         % colors
\usepackage{times}
\usepackage{epsfig}
\usepackage{graphicx}
\usepackage{float}
\usepackage{amsmath}
\usepackage{amssymb,bm}

\usepackage{caption}
\usepackage{subcaption}
\usepackage{array}

\usepackage{cleveref}
\usepackage{algorithm}
\usepackage{algpseudocode}
\usepackage{multirow}
\usepackage{tikz}
\usepackage{wrapfig,lipsum,booktabs}

\newcommand{\boldparagraph}[1]{\vspace{0.2cm}\noindent{\bf #1:}}

\title{MetaAvatar: Learning Animatable\\Clothed Human Models from Few Depth Images}
\author{%
  Shaofei Wang$^{1}$ \\ \texttt{shaofei.wang@inf.ethz.ch} \And Marko Mihajlovic$^{1}$ \\ \texttt{marko.mihajlovic@inf.ethz.ch} \And Qianli Ma$^{1,2}$ \\ \texttt{qianli.ma@tue.mpg.de } \And Andreas Geiger$^{2,3}$ \\ \texttt{a.geiger@uni-tuebingen.de} \And Siyu Tang$^{1}$ \\ \texttt{siyu.tang@inf.ethz.ch} \\
  \\
  $^1$ETH Z\"{u}rich \quad
  $^2$Max Planck Institute for Intelligent Systems, T\"{u}bingen \quad
  $^3$University of T\"{u}bingen\\
}

\begin{document}

\maketitle

\begin{abstract}
In this paper, we aim to create generalizable and controllable neural signed distance fields (SDFs) that represent clothed humans from monocular depth observations. Recent advances in deep learning, especially neural implicit representations, have enabled human shape reconstruction and controllable avatar generation from different sensor inputs. However, to generate realistic cloth deformations from novel input poses, watertight meshes or dense full-body scans are usually needed as inputs. Furthermore, due to the difficulty of effectively modeling pose-dependent cloth deformations for diverse body shapes and cloth types, existing approaches resort to per-subject/cloth-type optimization from scratch, which is computationally expensive. In contrast, we propose an approach that can quickly generate realistic clothed human avatars, represented as controllable neural SDFs, given only monocular depth images. We achieve this by using meta-learning to learn an initialization of a hypernetwork that predicts the parameters of neural SDFs. The hypernetwork is conditioned on human poses and represents a clothed neural avatar that deforms non-rigidly according to the input poses. Meanwhile, it is meta-learned to effectively incorporate priors of diverse body shapes and cloth types and thus can be much faster to fine-tune, compared to models trained from scratch. We qualitatively and quantitatively show that our approach outperforms state-of-the-art approaches that require complete meshes as inputs while our approach requires only depth frames as inputs and runs orders of magnitudes faster. Furthermore, we demonstrate that our meta-learned hypernetwork is very robust, being the first to generate avatars with realistic dynamic cloth deformations given as few as 8 monocular depth frames.
\end{abstract}

\section{Introduction}
\label{sec:intro}
% 3D human shape reconstruction has been a persistent research topic for decades in computer vision and computer graphics. It has many important applications in sports, design industry, online shopping and entertainments. Traditional approaches often work only in controlled environment, and rely on expensive capture setups that are not accessible to the general public. Human shape reconstruction from in-the-wild sensor inputs remains a challenging task in the community, while creating animatble avatar with desirable physical properties (\eg\ pose-dependent deformations of body tissues and cloths) from such reconstructions is a even harder task.~\todo{lots of traditional papers we need to cite}
Representing clothed humans as neural implicit functions is a rising research topic in the computer vision community. Earlier works in this direction address geometric reconstruction of clothed humans from static monocular images~\cite{Saito_ICCV2019,Saito_CVPR2020,tong2020geo-pifu,li2020monocular}, RGBD videos~\cite{DoubleFusion_CVPR_2018,RobustFusion_ECCV_2020,UnstructuredFusion_PAMI_2020,Li2020portrait,li2021posefusion} or sparse point clouds~\cite{IFNet} as direct extensions of neural implicit functions for rigid objects~\cite{Occupancy_Networks,DeepSDF,chen2018implicit_decoder,Michalkiewicz_2019_ICCV}. 
More recent works advocate to learn shapes in a canonical pose~\cite{ARCH_CVPR_2020,Bhatnagar_ECCV2020,PTF:CVPR:2021} in order to not only handle reconstruction, but also build controllable neural avatars from sensor inputs. However, these works do not model pose-dependent cloth deformation, limiting their realism. 

On the other hand, traditional parametric human body models~\cite{SMPL:2015,SMPL-X:2019,Xu_2020_CVPR,STAR:ECCV:2020} can represent pose-dependent soft tissue deformations of minimally-clothed human bodies. 
Several recent methods~\cite{Deng_ECCV2020,LEAP:CVPR:21} proposed to learn neural implicit functions to approximate such parametric models from watertight meshes. 
However, they cannot be straightforwardly extended to model clothed humans. 
SCANimate~\cite{SCANimate:CVPR:21} proposed to learn canonicalized dynamic neural Signed Distance Fields (SDFs) controlled by human pose inputs and trained with Implicit Geometric Regularization (IGR~\cite{Gropp:2020:ICML}), thus circumventing the requirement of watertight meshes. 
However, SCANimate works only on dense full-body scans with accurate surface normals and further requires expensive per-subject/cloth-type training.
These factors limit the applicability of SCANimate for building personalized human avatars from commodity RGBD sensors. 

% SCANimate~\cite{SCANimate:CVPR:21} proposed a weakly-supervised approach to canonicalize 3D scans of clothed humans and learn dynamic neural signed distance fields (SDFs) in the canonical space, which can vary with input human poses. These neural SDFs are trained with implicit geometric regularization~\cite{Gropp:2020:ICML}, circumventing the requirement of watertight meshes. However, SCANimate only works on dense, full body scans with accurate surface normals. It further requires per-subject/cloth-type training from scratch, which takes roughly a day to converge for each such model. These factors limit SCANimate's applicability in real-world scenarios such as building personalized avatars on portable devices.

\begin{figure}[t]
\centering
  \includegraphics[width=1.0\textwidth]{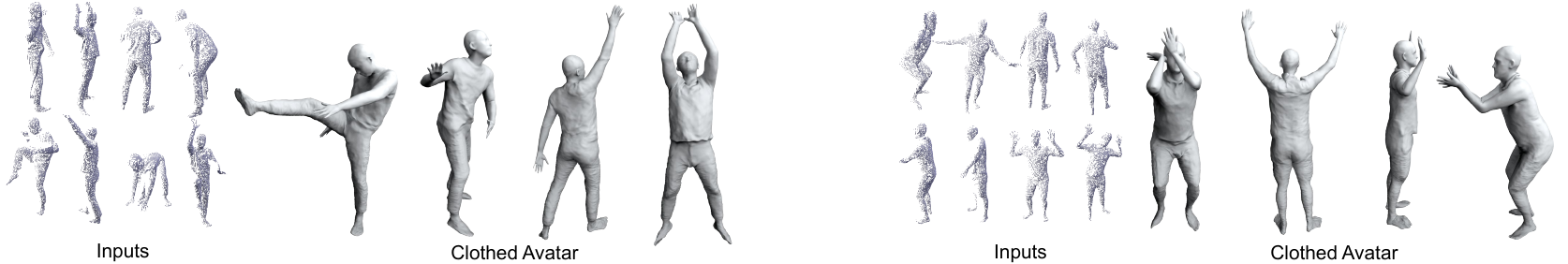}
  \caption{Given as few as 8 monocular depth images and their SMPL fittings, our meta-learned model yields a controllable neural SDF in 2 minutes which synthesizes realistic cloth deformations for unseen body poses. Here we show results of two different subjects wearing different clothes.}
  \label{fig:teaser}
\end{figure}

Contrary to all the aforementioned works, we propose to use meta-learning to effectively incorporate priors of dynamic neural SDFs of clothed humans, thus enabling fast fine-tuning (few minutes) for generating new avatars given only a few monocular depth images of unseen clothed humans as inputs. More specifically, we build upon recently proposed ideas of meta-learned initialization for implicit representations~\cite{sitzmann2019metasdf,tancik2020meta} to enable fast fine-tuning. Similar to~\cite{sitzmann2019metasdf}, we represent a specific category of objects (in our case, clothed human bodies in the canonical pose) with a neural implicit function and use meta-learning algorithms such as~\cite{MAML:ICML:2017,Reptile:arXiv:2018} to learn a meta-model. However, unlike~\cite{sitzmann2019metasdf,tancik2020meta}, where the implicit functions are designed for static reconstruction, we target the generation of dynamic neural SDFs that are \emph{controllable} by user-specified body poses. We observe that directly conditioning neural implicit functions (represented as a multi-layer perceptron) on body poses lacks the expressiveness to capture high-frequency details of diverse cloth types, and hence propose to meta-learn a hypernetwork~\cite{Ha:2017:ICLR} that predicts the parameters of the neural implicit function. Overall, the proposed approach, which we name \textit{MetaAvatar}, yields controllable neural SDFs with dynamic surfaces in minutes via fast fine-tuning, given only a few depth observations of an unseen clothed human and the underlying SMPL~\cite{SMPL:2015} fittings (Fig.~\ref{fig:teaser}) as inputs. Code and data are public at \href{https://neuralbodies.github.io/metavatar/}{\color{black}{https://neuralbodies.github.io/metavatar/}}.

% We structure the remainder of our paper as follows: in Section~\ref{sec:related} we review related works. In Section~\ref{sec:fundamentals} we introduce the recently proposed implicit skinning nets~\cite{SCANimate:CVPR:21}, which can be used to canonicalized 3D scans in a weakly-supervised manner as in~\cite{SCANimate:CVPR:21}. In Section~\ref{sec:meta_vatar}, we introduce the proposed MetaAvatar model, which extends the idea of metat-learning initialization of neural implicit functions~\cite{sitzmann2019metasdf,tancik2020meta} to meta-learning initialization of hypernetworks that predicts parameters of neural implicit functions. In Section~\ref{sec:exp} we quantitatively compare our MetaAvatar to two baselines, CAPE~\cite{CAPE:CVPR:20} and NASA~\cite{Deng_ECCV2020}, and also conduct ablation studies to confirm the necessity and effectiveness of our architectural design. In Section~\ref{sec:limits} we discuss limitations of our approach. In Section~\ref{sec:conclusion} we conclude and discuss possible future extensions.
% \begin{figure}
%     \centering
%     % \includegraphics[width=1.2\textwidth]{figures/MetaAvatar-overview.pdf}
%     \makebox[\textwidth][c]{\includegraphics[width=1.3\textwidth]{figures/MetaAvatar-overview.pdf}}%
%     \caption{\textbf{MetaAvatar overview.} \todo{Description}}
%     \label{fig:pipeline_overview}
% \end{figure}

% \begin{figure}
%     \begin{tikzpicture}
%     \node(a) {\makebox[\textwidth][l]{\includegraphics[width=0.7\textwidth]{figures/MetaAvatar-overview_v2.pdf}}};
%     \node at (a.east)
%     [
%     anchor=east,
%     xshift=0mm,
%     yshift=0mm
%     ]
%     {
%         \includegraphics[width=0.3\textwidth]{figures/hyper-avatar.pdf}
%     };
%     \end{tikzpicture}
%     \caption{\textbf{Overview of the proposed approach}. We meta-learn dynamic neural SDFs, represented as a meta-learned hypernetwork (Section~\ref{sec:meta_vatar}), that can effectively encode a unified prior of diverse body-shapes and cloth-types. At test-time, given depth images of clothed humans and the corresponding minimally-clothed SMPL registrations on unseen subjects wearing seen clothes (top row) or unseen clothes (bottom row), the meta-learned hypernetwork can be fast fine-tuned (often in minutes) to produce subject-specific dynamic neural SDFs.}
%     \label{fig:pipeline_overview}
% \end{figure}
\section{Related Work}
\label{sec:related}
Our approach lies at the intersection of clothed human body modeling, neural implicit representations, and meta-learning. We review related works in the following.

\boldparagraph{Clothed Human Body Modeling} Earlier works for clothed human body modeling utilize parametric human body models~\cite{SCAPE, Hasler2009CGF, Joo_2018_CVPR, SMPL:2015,SMPL-X:2019,Xu_2020_CVPR,STAR:ECCV:2020} combined with deformation layers~\cite{alldieck2019learning,alldieck2018video,Bhatnagar_ECCV2020,bhatnagar2020loopreg} to model cloth deformations. However, these approaches cannot model fine clothing details due to their fixed topology, and they cannot handle pose-dependent cloth deformations. 
Mesh-based approaches that handle articulated deformations of clothes either require accurate surface registration~\cite{deepwrinkles:2018:ECCV,CAPE:CVPR:20,Yang:ECCV:18,BUFF} or synthetic data~\cite{Guan12drape:dressing,gundogdu19garnet,patel20tailornet} for training.
Such requirement for data can be freed by using neural implicit surfaces~\cite{chen2021snarf,SCANimate:CVPR:21,tiwari21neuralgif,NPM:ICCV:2021}. 
For example, SCANimate~\cite{SCANimate:CVPR:21} proposed a weakly supervised approach to learn dynamic clothed human body models from 3D full-body scans which only requires minimally-clothed body registration. 
However, its training process usually takes one day for each subject/cloth-type combination and requires accurate surface normal information extracted from dense scans.
% , while our approach enables learning of clothed body models in minutes from as few as 8 depth images. 
Recent explicit clothed human models~\cite{SCALE:CVPR:21,POP:ICCV:2021,Zakharkin_2021_ICCV,DSFN:ICCV:2021} can also be learned from unregistered data. Like our method, concurrent work~\cite{POP:ICCV:2021} also models pose-dependent shapes across different subjects/cloth-types, but it requires full-body scans for training.
In contrast, our approach enables learning of clothed body models in minutes from as few as 8 depth images. 

\boldparagraph{Neural Implicit Representations} Neural implicit representations~\cite{chen2018implicit_decoder,Occupancy_Networks,Michalkiewicz_2019_ICCV, DeepSDF,Peng2021SAP} have been used to tackle both image-based~\cite{ARCH_CVPR_2020,li2020monocular,tong2020geo-pifu,Saito_ICCV2019,Saito_CVPR2020,raj2020anr,peng2020neural,pamir2020} and point cloud-based~\cite{Bhatnagar_ECCV2020, IFNet} clothed human reconstruction. Among these works, ARCH~\cite{ARCH_CVPR_2020} was the first one to represent clothed human bodies as a neural implicit function in a canonical pose. However, ARCH does not handle pose-dependent cloth deformations. Most recently, SCANimate~\cite{SCANimate:CVPR:21} proposed to condition neural implicit functions on joint-rotation vectors (in the form of unit quaternions), such that the canonicalized shapes of the neural avatars change according to the joint angles of the human body, thus representing pose-dependent cloth deformations. However, diverse and complex cloth deformations make it hard to learn a unified prior from different body shapes and cloth types, thus SCANimate resorts to per-subject/cloth-type training which is computationally expensive.   

\boldparagraph{Meta-Learning}
Meta-learning is typically used to address few-shot learning, where a few training examples of a new task are given, and the model is required to learn from these examples to achieve good performance on the new task~\cite{snell2017prototypical,Perez-Rua_2020_CVPR,kang2019few,Gui_2018_ECCV,pmlr-v87-alet18a,dong2018fewshot,shaban2017oneshot,Wang_2019_ICCV,rakelly2018fewshot,reed2018fewshot,gordon2019metalearning,eslami2018neural,Zakharov_2019_ICCV,wang2018fewshotvid2vid,garnelo2018neural,garnelo2018conditional,kim2019attentive,MetaHyper:NeurIPSW:2020}. We focus on optimization-based meta-learning, where Model-Agnostic Meta Learning (MAML~\cite{MAML:ICML:2017}), Reptile~\cite{Reptile:arXiv:2018} and related alternatives are typically used to learn such models~\cite{MetaSGD:arXiv:2017,trainMAML:ICLR:2019,learn2learn:NeurIPS:2016,optimAsFewshot:ICLR:2017,learn2optim:ICLR:2017,genearlizedInner:arXiv:2019}. In general, this line of algorithms tries to learn a "meta-model" that can be updated quickly from new observations with only few gradient steps. Recently, meta-learning has been used to learn a universal initialization of implicit representations for static neural SDFs~\cite{sitzmann2019metasdf} and radiance fields~\cite{tancik2020meta}. MetaSDF~\cite{sitzmann2019metasdf} demonstrates that only a few gradient update steps are needed to achieve comparable or better results than slower auto-decoder-based approaches~\cite{DeepSDF}. However,~\cite{sitzmann2019metasdf,tancik2020meta} only meta-learn static representations, whereas we are interested in dynamic representations conditioned on human body poses. To our best knowledge, we are the first to meta-learn the hypernetwork to generate the parameters of neural SDF networks.
\section{Fundamentals}
\label{sec:fundamentals}
We start by briefly reviewing the linear blend skinning (LBS) method \cite{SMPL:2015} and the recent implicit skinning networks~\cite{LEAP:CVPR:21,SCANimate:CVPR:21} that learn to predict skinning weights of cloth surfaces in a weakly supervised manner. Using the learned implicit skinning networks allows us to canonicalize meshes or depth observations of clothed humans, given only minimally-clothed human body model registrations to the meshes. Canonicalization of meshes or points is a necessary step as the dynamic neural SDFs introduced in Section~\ref{sec:meta_vatar} are modeled in canonical space. 

\subsection{Linear Blend Skinning}
\label{sec:LBS}
Linear blend skinning (LBS) is a commonly used technique to deform parametric human body models~\cite{SCAPE, Hasler2009CGF, SMPL:2015, SMPL-X:2019, STAR:ECCV:2020, Xu_2020_CVPR} according to user-specified rigid bone transformations. Given a set of $N$ points in a canonical space,  $\hat{\mathbf{X}} = \{ \hat{\mathbf{x}}^{(i)} \}_{i=1}^{N}$, LBS takes a set of rigid bone transformations (in our case we use 23 local transformations plus one global transformation, assuming an underlying SMPL model) $\{ \mathbf{B}_b \}_{b=1}^{24}$ as inputs, each $\mathbf{B}_b$ being a $4 \times 4$ rotation-translation matrix. For a 3D point $\hat{\mathbf{x}}^{(i)} \in \hat{\mathbf{X}}$~\footnote{with slight abuse of notation, we also use $\hat{\mathbf{x}}$ to represent points in homogeneous coordinates when necessary.}, a skinning weight vector is a probability simplex $\mathbf{w}^{(i)} \in [ 0, 1 ]^{24}, \text{s.t.} \sum_{b=1}^{24} \mathbf{w}_b^{(i)} = 1$, that defines the affinity of the point $\hat{\mathbf{x}}^{(i)}$ to each of the bone transformations $\{ \mathbf{B}_b \}_{b=1}^{24}$. The set of transformed points $\mathbf{X} = \{ \mathbf{x}^{(i)} \}_{i=1}^{N}$ of the clothed human is related to $\hat{\mathbf{X}}$ via:
\begin{align}
    \label{eqn:LBS}
    \mathbf{x}^{(i)} &= LBS\left(\hat{\mathbf{x}}^{(i)}, \{ \mathbf{B}_b \}, \mathbf{w}^{(i)}\right) = \left(\sum_{b=1}^{24} \mathbf{w}^{(i)}_{b} \mathbf{B}_b\right) \hat{\mathbf{x}}^{(i)}, \quad \forall i = 1, \ldots, N \\ 
    \label{eqn:inverse-LBS}
    \hat{\mathbf{x}}^{(i)} &= LBS^{-1}\left(\mathbf{x}^{(i)}, \{ \mathbf{B}_b \}, \mathbf{w}^{(i)}\right) = \left(\sum_{b=1}^{24} \mathbf{w}^{(i)}_{b} \mathbf{B}_b\right)^{-1} \mathbf{x}^{(i)}, \quad \forall i = 1, \ldots, N
\end{align}
where Eq.~\eqref{eqn:LBS} is referred to as the LBS function and Eq.~\eqref{eqn:inverse-LBS} is referred to as the inverse-LBS function. The process of applying Eq.~\eqref{eqn:LBS} to all points in $\hat{\mathbf{X}}$ is often referred to as \textit{forward skinning} while the process of applying Eq.~\eqref{eqn:inverse-LBS} is referred to as \textit{inverse skinning}.

\subsection{Implicit Skinning Networks}
\label{sec:lbs_net}
Recent articulated implicit representations \cite{LEAP:CVPR:21, SCANimate:CVPR:21} have proposed to learn functions that predict the forward/inverse skinning weights for arbitrary points in $\mathbb{R}^3$. We follow this approach, but take advantage of a convolutional point-cloud encoder~\cite{ConvONet} for improved generalization. Formally, we define the implicit forward and inverse skinning networks as $h_{\text{fwd}}(\cdot, \cdot): (\mathbb{R}^{3 \times K}, \mathbb{R}^3) \mapsto \mathbb{R}^{24}$ and $h_{\text{inv}}(\cdot, \cdot): (\mathbb{R}^{3 \times K}, \mathbb{R}^3) \mapsto \mathbb{R}^{24}$, respectively. Both networks take as input a point cloud with $K$ points and a query point for which they predict skinning weights. Therefore, we can analogously re-define Eq.~(\ref{eqn:LBS}, \ref{eqn:inverse-LBS}) respectively as:
\begin{align}
    \label{eqn:LBS-with-skinning-net}
    \mathbf{x}^{(i)} &= \left(\sum_{b=1}^{24} h_{\text{fwd}}(\hat{\mathbf{X}}, \hat{\mathbf{x}}^{(i)})_{b} \mathbf{B}_b\right) \hat{\mathbf{x}}^{(i)}, \quad \forall i = 1, \ldots, N \\ 
    \label{eqn:inverse-LBS-with-skinning-net}
    \hat{\mathbf{x}}^{(i)} &= \left(\sum_{b=1}^{24} h_{\text{inv}}(\mathbf{X}, \mathbf{x}^{(i)})_{b} \mathbf{B}_b\right)^{-1} \mathbf{x}^{(i)}, \quad \forall i = 1, \ldots, N
\end{align}
\boldparagraph{Training the Skinning Network} We follow the setting of SCANimate~\cite{SCANimate:CVPR:21}, where a dataset of observed point clouds $\{ \mathbf{X} \}$ and their underlying SMPL registration are known. For a sample $\mathbf{X}$ in the dataset, we first define the re-projected points $\bar{\mathbf{X}} = \{ \bar{\mathbf{x}} \}_{i=1}^{N}$ as $\mathbf{X}$ mapped to canonical space via Eq.~\eqref{eqn:inverse-LBS-with-skinning-net} and then mapped back to transformed space via Eq.~\eqref{eqn:LBS-with-skinning-net}. We then define the training loss:
\begin{align}
\mathcal{L} (\mathbf{X}) = \lambda_{r} \mathcal{L}_{r} + \lambda_{s} \mathcal{L}_{s} + \lambda_{skin} \mathcal{L}_{skin} \,,
\end{align}
where $\mathcal{L}_r$ represents a re-projection loss that penalizes the L2 distance between an input point $\mathbf{x}$ and the re-projected point $\bar{\mathbf{x}}$, $\mathcal{L}_s$ represents L1 distances between the predicted forward skinning weights and inverse skinning weights, and $\mathcal{L}_{skin}$ represents the L1 distances between the predicted (forward and inverse) skinning weights and the barycentrically interpolated skinning weights $\mathbf{w}^{(i)}$ on the registered SMPL shape that is closest to point $\mathbf{x}^{(i)}$; please refer to Appendix~\ref{appx:skinning_nets} for hyperparameters and details. 

We train two skinning network types, the first one takes a partial point cloud extracted from a depth image as input and performs the \textit{inverse skinning}, while the second one takes a full point cloud sampled from iso-surface points generated from the dynamic neural SDF in the canonical space and performs \textit{forward skinning}.

\boldparagraph{Canonicalization} We use the learned inverse skinning network to canonicalize complete or partial point clouds $\{ \hat{\mathbf{X}} \}$ via Eq.~\eqref{eqn:inverse-LBS-with-skinning-net} which are further used to learn the canonicalized dynamic neural SDFs. 
\section{MetaAvatar}
\label{sec:meta_vatar}
% In this section, we introduce our novel MetaAvatar representation. 
Our approach meta-learns a unified clothing deformation prior from the training set that consists of different subjects wearing different clothes. 
This meta-learned model is further efficiently fine-tuned to produce a dynamic neural SDF from an arbitrary amount of fine-tuning data of unseen subjects. In extreme cases, MetaAvatar requires as few as 8 depth frames and takes only 2 minutes for fine-tuning to yield a subject/cloth-type-specific dynamic neural SDF (Fig.~\ref{fig:teaser}). 

We assume that each subject/cloth-type combination in the training set has a set of registered bone transformations and canonicalized points, denoted as $\{\{\mathbf{B}_b\}_{b=1}^{24}, \hat{\mathbf{X}} \}$. Points in $\hat{\mathbf{X}}$ are normalized to the range $[-1, 1]^3$ according to their corresponding registered SMPL shape. With slight abuse of notation, we also define $\mathbf{X}$ as all possible points in $[-1, 1]^3$. % A naive solution would be fitting neural SDF $f(\mathbf{x}, \mathbf{c}_{\mathbf{B}}): ([-1, 1]^3, \mathbb{R}^{|\mathbf{c}_{\mathbf{B}}|}) \xrightarrow[]{} \mathbb{R}$ directly to canonicalized points $\{ \hat{\mathbf{X}} \}$,
Our goal is to meta-learn a hypernetwork~\cite{Ha:2017:ICLR,sitzmann2019srns} which takes $\{ \mathbf{B}_b \}_{b=1}^{24}$ ($\{ \mathbf{B}_b \}$ for shorthand) as inputs and predicts \textit{parameters of the neural SDFs} in the canonical space. Denoting the hypernetwork as $g_{\psi}(\{ \mathbf{B}_b \})$ and the predicted neural SDF as $f_{\phi}(\mathbf{x}) |_{\phi = g_{\psi}(\{ \mathbf{B}_b \})}$, we use the following IGR~\cite{Gropp:2020:ICML} loss to supervise the learning of $g$: 
% We meta-learn a hypernetwork that where $\mathbf{c}_{\mathbf{B}}$ is some feature extracted from rigid bone-transformations $\{\mathbf{B}_b\}_{b=1}^{24}$. In the Section~\ref{sec:exp} we will ablate on different choices of $\mathbf{c}_{\mathbf{B}}$ including unit quaternionis of relative joint-rations~\cite{SCANimate:CVPR:21} and hierarchical structural encoder~\cite{LEAP:CVPR:21}. Learning $f$ can be done by optimizing the following implicit geometric regularization (IGR~\cite{Gropp:2020:ICML}) loss:
%
\begin{align}
\label{eqn:IGR}
    \mathcal{L}_{\text{IGR}} (f_{\phi} (\hat{\mathbf{X}})|_{\phi = g_{\psi}(\{ \mathbf{B}_b \})}) =& \sum_{\mathbf{x} \in \hat{\mathbf{\mathbf{X}}}} \lambda_{sdf} \Big| f_{\phi}(\mathbf{x}) |_{\phi = g_{\psi}(\{ \mathbf{B}_b \})} \Big| + \lambda_{\mathbf{n}} \left(1 - \langle \mathbf{n} (\mathbf{x}), \nabla_{\mathbf{x}} f_{\phi}(\mathbf{x}) |_{\phi = g_{\psi}(\{ \mathbf{B}_b \})}\rangle\right) \nonumber \\
    & \qquad + \lambda_{E} \Big| \lVert \nabla_{\mathbf{x}} f_{\phi}(\mathbf{x}) |_{\phi = g_{\psi}(\{ \mathbf{B}_b \})} \rVert_2 - 1 \Big| \nonumber \qquad \qquad \text{(on-surface loss)} \\
    &+ \sum_{\mathbf{x} \sim \mathbf{X} \setminus \hat{\mathbf{X}}} \lambda_{O} \exp \left(-\alpha \cdot \Big|f_{\phi}(\mathbf{x}) |_{\phi = g_{\psi}(\{ \mathbf{B}_b \})} \Big|\right) \nonumber\\
    & \qquad \qquad + \lambda_{E} \Big| \lVert \nabla_{\mathbf{x}} f_{\phi}(\mathbf{x}) |_{\phi = g_{\psi}(\{ \mathbf{B}_b \})} \rVert_2 - 1 \Big| \qquad \text{(off-surface loss)}
\end{align}
where $\mathbf{n}_{\mathbf{x}}$ is the surface normal of point $\mathbf{x}$. We assume that this information, along with the ground-truth correspondences from transformed space to canonical space is available when learning the meta-model on the training set, but do not require this for fine-tuning $g$ on unseen subjects. 
%As we will demonstrate in the experiment section, if we directly meta-learn $f$ using the loss in Eq.~\eqref{eqn:IGR}, then the models fine-tuned from this meta-learned function $f$ lacks expressiveness and gives overly smooth surface reconstruction results. 

% In practice, we decompose the meta-learning of the hypernetwork into two steps: \textbf{1)} we learn a meta-SDF~\cite{sitzmann2019metasdf}, without conditioning on $\{ \mathbf{B}_b \}$; this meta-SDF can quickly adapt to a single depth input with only a few gradient steps, \textbf{2)} we meta-learn a hypernetwork that takes $\{ \mathbf{B}_b \}$ as input and predicts the \textit{residuals} to the parameters of the previously learned meta-SDF.

In practice, we found that directly learning the hypernetwork $g_{\psi}$ via Eq.~\eqref{eqn:IGR} does not converge, and thus we decompose the meta-learning of $g_{\psi}$ into two steps. First, we learn a meta-SDF~\cite{sitzmann2019metasdf} (without conditioning on $\{ \mathbf{B}_b \}$, Sec.~\ref{sec:meta-sdf}), and then we meta-learn a hypernetwork that takes $\{ \mathbf{B}_b \}$ as input and predicts the \textit{residuals} to the parameters of the previously learned meta-SDF (Sec.~\ref{sec:meta-hyper-sdf}).

% We meta-learn a hypernetwork~\cite{Ha:2017:ICLR,sitzmann2019srns} which takes $\mathbf{c}_{\mathbf{B}}$ as inputs and outputs \textit{parameters of neural SDFs} in the canonical space. We decompose the meta-learning of the hypernetwork into two steps: 1) we learn a meta-SDF~\cite{sitzmann2019metasdf} without conditioning on $\bm{\theta}$ that can quickly adapt to a single depth input with only a few gradient steps 2) we meta-learn a hypernetwork that takes $\bm{\theta}$ as input and predicts the \textit{residuals} to the parameters of the previously learned meta-SDF.

\subsection{Meta-learned Initialization of Static Neural SDFs} \label{sec:meta-sdf}
\begin{wrapfigure}{r}{0.5\textwidth}
    \includegraphics[width=0.5\textwidth]{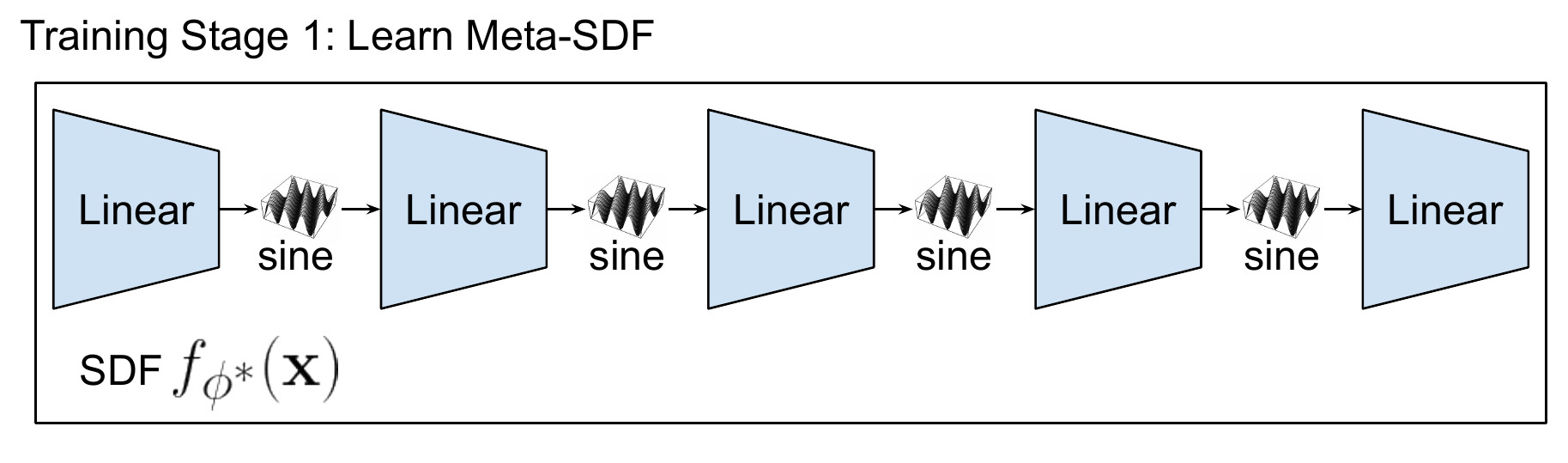}
    \caption{\textbf{Overview of the meta-SDF network}. We use a 5-layer SIREN~\cite{sitzmann2019metasdf} network with 256 neurons for each layer.}
    \label{fig:pipeline_stage1}
\end{wrapfigure}
To effectively learn a statistical prior of clothed human bodies, we ignore the input bone transformations $\{ \mathbf{B}_b \}$ and meta-learn the static neural SDF $f_{\phi}(\mathbf{x}): [-1, 1]^3 \mapsto \mathbb{R}$, parameterized by $\phi$, from all canonicalized points of subjects \textit{with different genders, body shapes, cloth types, and poses}. 
Furthermore, for faster and more stable convergence, the neural SDF $f_{\phi}$ function additionally leverages the periodic activation functions~\cite{sitzmann2019siren}. 
%The network architecture is shown in Fig.~\ref{fig:pipeline_stage1}.

The full meta-learning algorithm for the static neural SDFs is described in Alg.~\ref{alg:reptile_sdf}.

\begin{algorithm}
%\scriptsize
\caption{Meta-learning SDF with Reptile~\cite{Reptile:arXiv:2018}}
\begin{algorithmic}[1] 
  \Statex \textbf{Initialize}: meta-network parameters $\phi$, meta learning rate $\beta$, inner learning rate $\alpha$, max training iteration $N$, inner-loop iteration $m$, batch size $M$
  \For{$i = 1, \ldots, N$}
    \State Sample a batch of $M$ training samples $\{ \hat{\mathbf{X}}^{(j)} \}_{j=1}^{M}$
    \For {$j = 1, \ldots, M$}
      \State $\phi^{(j)}_0 = \phi$
      \For{$k = 1, \ldots, m$}
        \State $\phi^{(j)}_{k} = \phi^{(j)}_{k-1} - \alpha \nabla_{\phi}\mathcal{L}_{\text{IGR}}(f_{\phi}(\hat{\mathbf{X}}^{(j)})|_{\phi=\phi^{(j)}_{k-1}})$
      \EndFor
    \EndFor
    \State $\phi \leftarrow \phi + \beta \frac{1}{M} \sum_{j=1}^{M} (\phi^{(j)}_m - \phi)$
  \EndFor
\end{algorithmic}
\label{alg:reptile_sdf}
\end{algorithm}

\subsection{Meta-learned Initialization of HyperNetwork for Dynamic Neural SDFs}
\label{sec:meta-hyper-sdf}
The meta-learned static neural SDF explained in the previous section can efficiently adapt to new observations, however it is not controllable by user-specified bone transformations $\{ \mathbf{B}_b \}$. Therefore, to enable non-rigid pose-dependent cloth deformations, we further meta-learn a hypernetwork~\cite{Ha:2017:ICLR} to predict \textit{residuals} to the learned parameters of the meta-SDF in Alg.~\ref{alg:reptile_sdf}. 

The key motivation for meta-learning the hypernetwork is to build an effective unified prior for articulated clothed humans,  
which enables the recovery of the non-rigid clothing deformations at test time via the efficient fine-tuning process from several depth images of unseen subjects. 
\begin{algorithm*}
%\scriptsize
\caption{Meta-learning hypernetwork with Modified Reptile}
\begin{algorithmic}[1] 
  \Statex \textbf{Initialize}: meta-hypernetwork parameters $\psi$, pre-trained meta-SDF parameters $\phi^*$, meta learning rate $\beta$, inner learning rate $\alpha$, max training iteration $N$, inner-loop iteration $m$.
  \For{$i = 1, \ldots, N$}
    \State $\psi_0 = \psi$
    \State Randomly choose a subject/cloth-type combination $n$
    \State Uniformly sample $M \sim \{1, \ldots, D^{(n)}\}$ where $D^{(n)}$ is the number of datapoints of subject/cloth-type combination $n$
    \State Sample $M$ datapoints from subject/cloth-type combination $n$, denoting these datapoints as $\mathcal{S} = \{ \{ \mathbf{B}_b \}^{(j)}, \hat{\mathbf{X}}^{(j)} \}_{j=1}^{M}$ 
    \For{$k = 1, \ldots, m$}
      \State $\mathcal{L} = \frac{1}{M} \sum_{(\{ \mathbf{B}_b \}, \hat{\mathbf{X}}) \in \mathcal{S}} \mathcal{L}_{\text{IGR}}(f_{\phi}(\hat{\mathbf{X}})|_{\phi=g_{\psi_{k-1}}(\{ \mathbf{B}_b \}) + \phi^*})$ \label{alg:hyper-meta-loss}
      \State $\psi_{k} = \psi_{k-1} - \alpha \nabla_{\psi_{k-1}} \mathcal{L}$ \label{alg:hyper-meta-inner}
    \EndFor
    \State $\psi \leftarrow \psi + \beta (\psi_{m} - \psi_0)$
  \EndFor
\end{algorithmic}
\label{alg:reptile_hyper-sdf}
\end{algorithm*}
\begin{wrapfigure}{r}{0.5\textwidth}
    \includegraphics[width=0.5\textwidth]{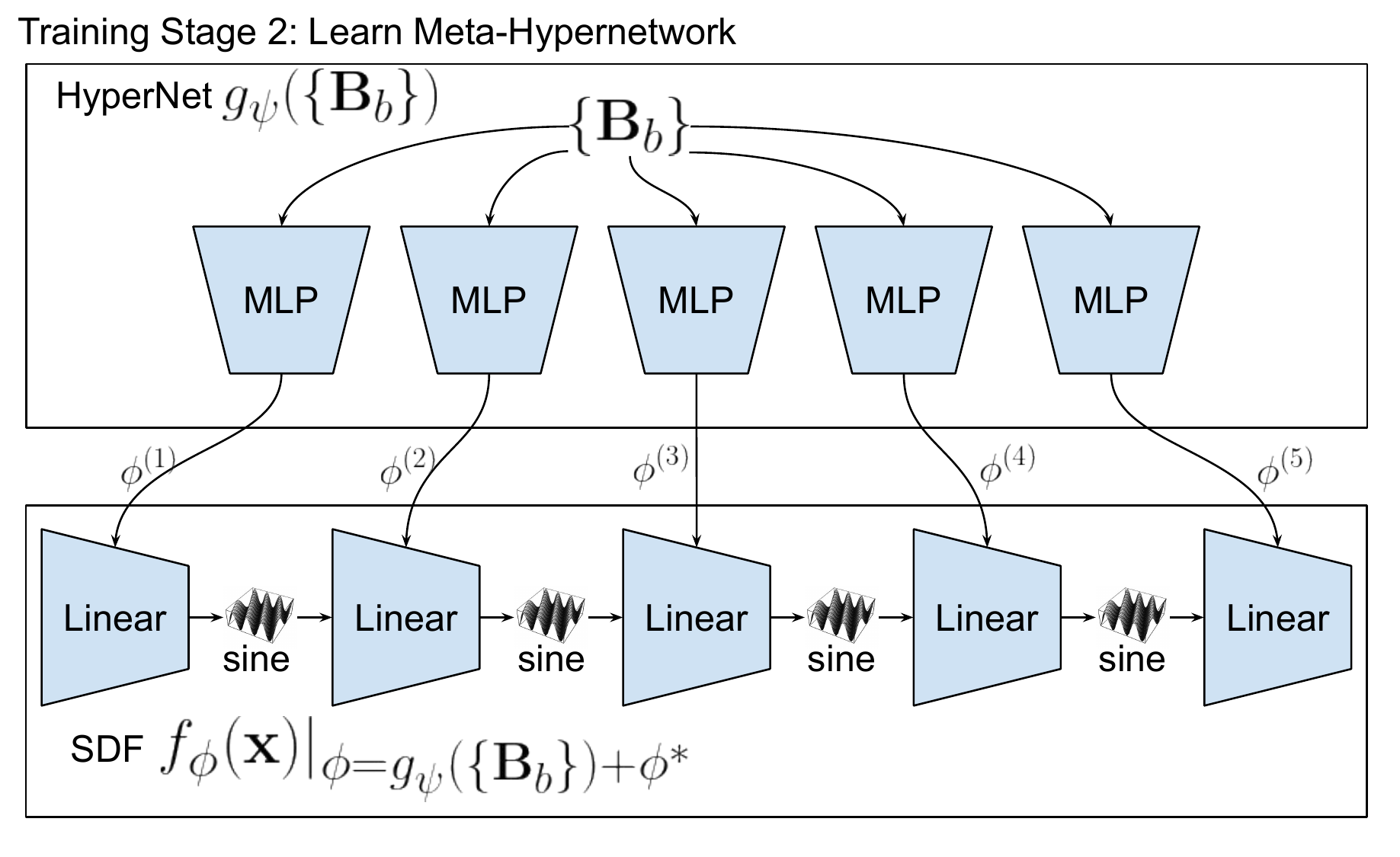}
    \caption{\textbf{Overview of the meta-hypernetwork}. It predicts residuals to $\phi^*$ which is learned in Sec.~\ref{sec:meta-sdf}}
    \label{fig:pipeline_stage2}
\end{wrapfigure}
Denoting the meta-SDF learned by Alg.~\ref{alg:reptile_sdf} as $\phi^*$ and our hypernetwork as $g_{\psi}(\{ \mathbf{B}_b \})$, we implement Alg.~\ref{alg:reptile_hyper-sdf}. This algorithm differs from the original Reptile~\cite{Reptile:arXiv:2018} algorithm in that it tries to optimize the inner-loop on arbitrary amount of data.
Note that for brevity the loss in the inner-loop (line~\ref{alg:hyper-meta-loss}-line~\ref{alg:hyper-meta-inner}) is computed over the whole batch $\mathcal{S}$, whereas in practice we used stochastic gradient descent (SGD) with fixed mini-batch size over $\mathcal{S}$ since $\mathcal{S}$ can contain hundreds of samples; SGD is used with the mini-batch size of 12 for the inner-loop.

\begin{figure}
    \includegraphics[width=\textwidth]{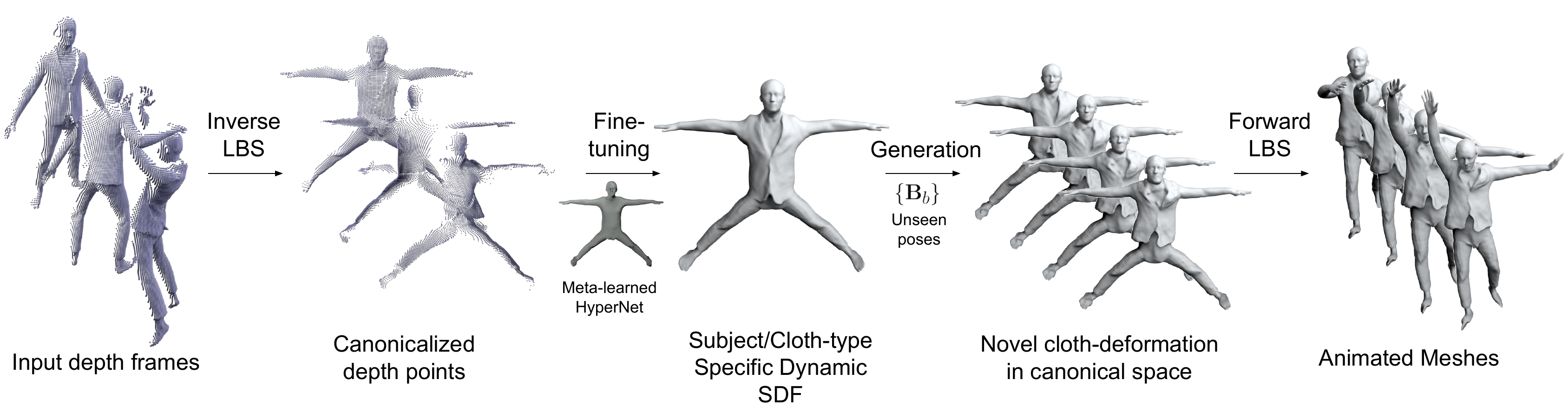}
    \caption{\textbf{Overview of our inference pipeline}. The inverse LBS net (Sec.~\ref{sec:lbs_net}) takes a small set of input depth frames together with their underlying SMPL registrations to canonicalize the depth points; then the meta-learned hypernetwork (Sec.~\ref{sec:meta-hyper-sdf}) is fine-tuned to represent the instance specific dynamic SDF; given novel poses, the updated hypernetwork generates pose-dependent cloth-deformations in canonical space, and the animated meshes are obtained via the forward LBS network (Sec.~\ref{sec:lbs_net}).}
    \label{fig:inference}
\end{figure}

\boldparagraph{Inference} At test-time, we are given a small fine-tuning set $\{ \{ \mathbf{B}_b \}^{fine, (j)}, \hat{\mathbf{X}}^{fine, (j)} \}_{j=1}^{M}$ and the validation set $\{ \{ \mathbf{B}_b \}^{val, (j)} \}_{j=1}^{K}$. 
The fine-tuning set is used to optimize the hypernetwork parameters $\psi$ ($m=256$ SGD epochs) that are then used to generate neural SDFs from bone transformations available in the validation set. 
The overall inference pipeline including the inverse and the forward LBS stages is shown in Fig.~\ref{fig:inference}.

\boldparagraph{Bone Transformation Encoding}
We found that a small hierarchical MLP proposed in LEAP~\cite{LEAP:CVPR:21} for encoding bone transformations works slightly better than the encoding of unit quaternions used in SCANimate~\cite{SCANimate:CVPR:21}. Thus, we employ the hierarchical MLP encoder to encode $\{ \mathbf{B}_b \}$ for $g$ unless specified otherwise;  we ablate different encoding types in the experiment section.
\section{Experiments}
\label{sec:exp}
We validate the proposed MetaAvatar model for learning meta-models and controllable dynamic neural SDFs of clothed humans %via fast fine-tuning.
by first comparing our MetaAvatar to the established approaches~\cite{Deng_ECCV2020,LEAP:CVPR:21,SCANimate:CVPR:21}. 
Then, we ablate the modeling choices for the proposed controllable neural SDFs. 
And lastly, we demonstrate MetaAvatar's capability to tackle the challenging task of learning animatable clothed human models from reduced data, to the point that only 8 depth images are available as input.

\boldparagraph{Datasets}
We use the CAPE dataset~\cite{CAPE:CVPR:20} as the major test bed for our experiments. This dataset consists of 148584 pairs of clothed meshes, capturing 15 human subjects wearing different clothes while performing different actions. We use 10 subjects for meta-learning, which we denote as the training set. We use four unseen subjects (00122, 00134, 00215, 03375)\footnote{We ignore subject 00159 because it has significantly less data compared to other subjects.} for fine-tuning and validation; for each of these four subjects, the corresponding action sequences are split into fine-tuning set and validation set. The fine-tuning set is used for fine-tuning %training/fine-tuning established baselines and 
the MetaAvatar models, it is also used to evaluate pose interpolation task. The validation set is used for evaluating novel pose extrapolation. Among the four unseen subjects, two of them (00122, 00215) perform actions that are present in the training set for the meta-learning; 
we randomly split actions of these two subjects with 70\% fine-tuning and 30\% validation. Subject 00134 and 03375 perform two trials of actions unseen in the training set for meta-learning. We use the first trial as the fine-tuning set and the second trial as the validation set. Subject 03375 also has one cloth type (blazer) that is unseen during meta-learning. 

\boldparagraph{Baselines}
We use NASA~\cite{Deng_ECCV2020}, LEAP~\cite{LEAP:CVPR:21}, and SCANimate~\cite{SCANimate:CVPR:21} as our baselines. NASA and SCANimate cannot handle multi-subject-cloth with a single model so we train per-subject/cloth-type models from scratch for each of them on the fine-tuning set. LEAP is a generalizable neural-implicit human body model that has shown to work on minimally-clothed bodies. We extend LEAP by adding a one-hot encoding to represent different cloth types (similarly to \cite{CAPE:CVPR:20}) and train it jointly on the full training and the fine-tuning set.

As for the input format, we use depth frames rendered from CAPE meshes for our MetaAvatar. To render the depth frames, we fixed the camera and rotate the CAPE meshes around the y-axis (in SMPL space) at different angles with an interval of 45 degrees; note that for each mesh we only render it on one angle, simulating a monocular camera taking a round-view of a moving person. For the baselines, we use watertight meshes and provide the occupancy~\cite{Occupancy_Networks} loss to supervise the training of NASA and LEAP, while sampling surface points and normals on watertight meshes to provide the IGR loss supervision for SCANimate. Note that our model is at great disadvantage, as for fine-tuning we only use discrete monocular depth observations without accurate surface normal information. 

\boldparagraph{Tasks and Evaluation}
Our goal is to generate realistic clothing deformations from arbitrary input human pose parameters. 
To systematically understand the generalization capability of the MetaAvatar representation, we validate the baselines and MetaAvatar on two tasks, pose interpolation and extrapolation. 
For interpolation, we sample every 10th frame on the fine-tuning set for training/fine-tuning, and sample every 5th frame (excluding the training frames) also on the fine-tuning set for validation. 
For extrapolation, we sample every 5th frame on the fine-tuning set for training, and sample every 5th frame on the validation set for validation. 

Interpolation is evaluated using three metrics: point-based ground-truth-to-mesh distance ($D_p \downarrow$, in cm), face-based ground-truth-to-mesh distance ($D_f \downarrow$, in cm), and point-based ground-truth-to-mesh normal consistency ($NC \uparrow$, in range $[-1, 1]$). For computing these interpolation metrics we ignore non-clothed body parts such as hands, feet, head, and neck. For extrapolation, we note that cloth-deformation are often stochastic; in such a case, predicting overly smooth surfaces can result in lower distances and higher normal consistency. Thus, we also conduct a large-scale perceptual study using Amazon Mechanical Turk, and report the perceptual scores (PS $\uparrow$) which reflects the percentage of users who favor the outputs of baselines over MetaAvatar. Details about user study design can be found in Appendix~\ref{appx:perceptual_study}.

\subsection{Evaluation Against Baselines}
\label{sec:exp_baselines}
In this section, we report results on both interpolation and extrapolation tasks against various baselines described above. For NASA and SCANimate, we train one model for each subject/cloth-type combination on the fine-tuning set. For them it usually takes several thousand of epochs to converge for each subject/cloth-type combination, which roughly equals to 10-24 hours of training. For LEAP, we train a single model on both the training and the fine-tuning set using two days. For the MetaAvatar, we meta-learn a single model on the training set, and for each subject/cloth-type combination we fine-tune the model for 256 epochs to produce subject/cloth-type specific models. The exact fine-tuning time ranges from 40 minutes to 3 hours depending on the amount of available data since we are running a fixed number of epochs; see Appendix~\ref{appx:timing} for detailed runtime comparison on each subject/cloth-type combination. Note that MetaAvatar uses \textit{partial} depth observations while the other baselines are trained on \textit{complete} meshes. The results are reported in Table~\ref{tab:eval_baselines}.

\begin{figure}[t]
\begin{minipage}[b]{0.48\textwidth}
\centering
  \begin{subfigure}{0.3\textwidth}
  \captionsetup{labelformat=empty, font=scriptsize}
    \includegraphics [trim=6cm 7cm 6cm 5cm, width=\textwidth]{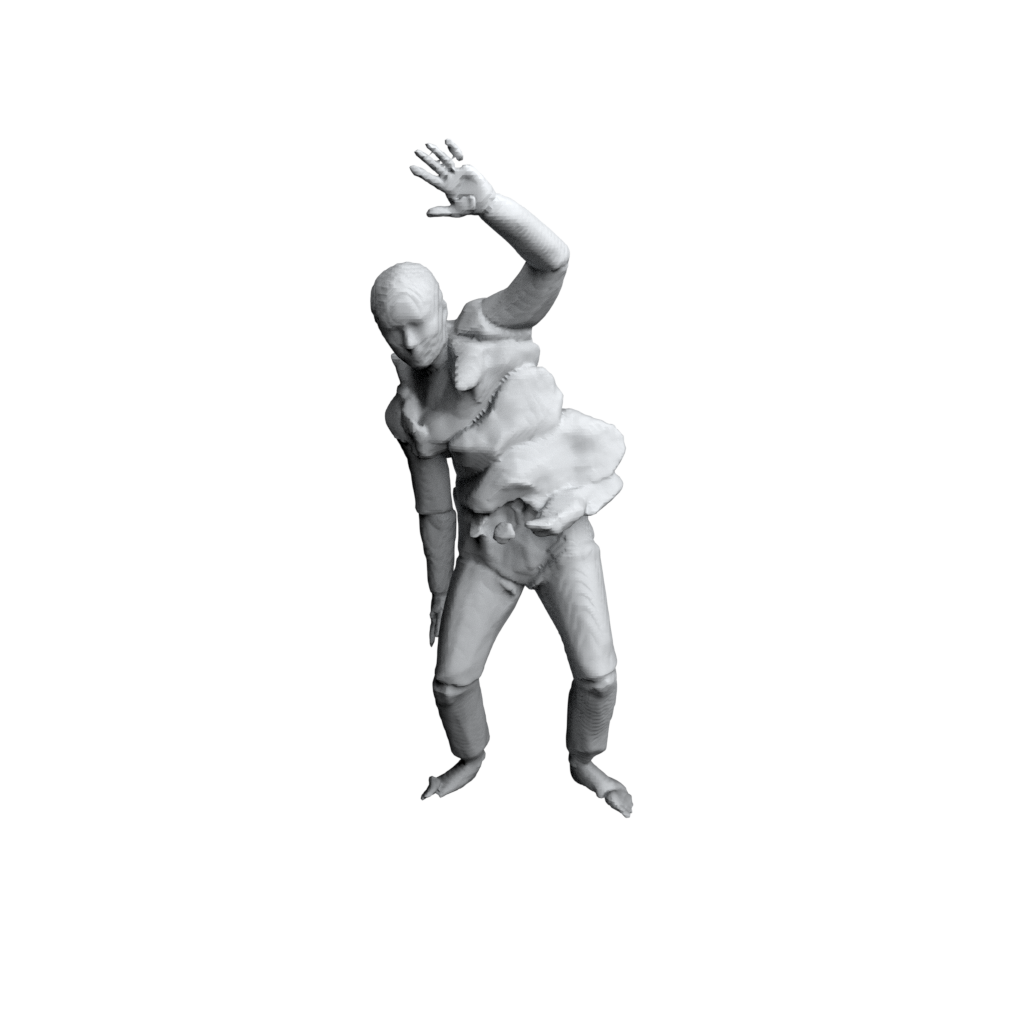}
  \caption{NASA}
  \end{subfigure}
  \begin{subfigure}{0.3\textwidth}
  \captionsetup{labelformat=empty, font=scriptsize}
    \includegraphics [trim=6cm 7cm 6cm 5cm, width=\textwidth]{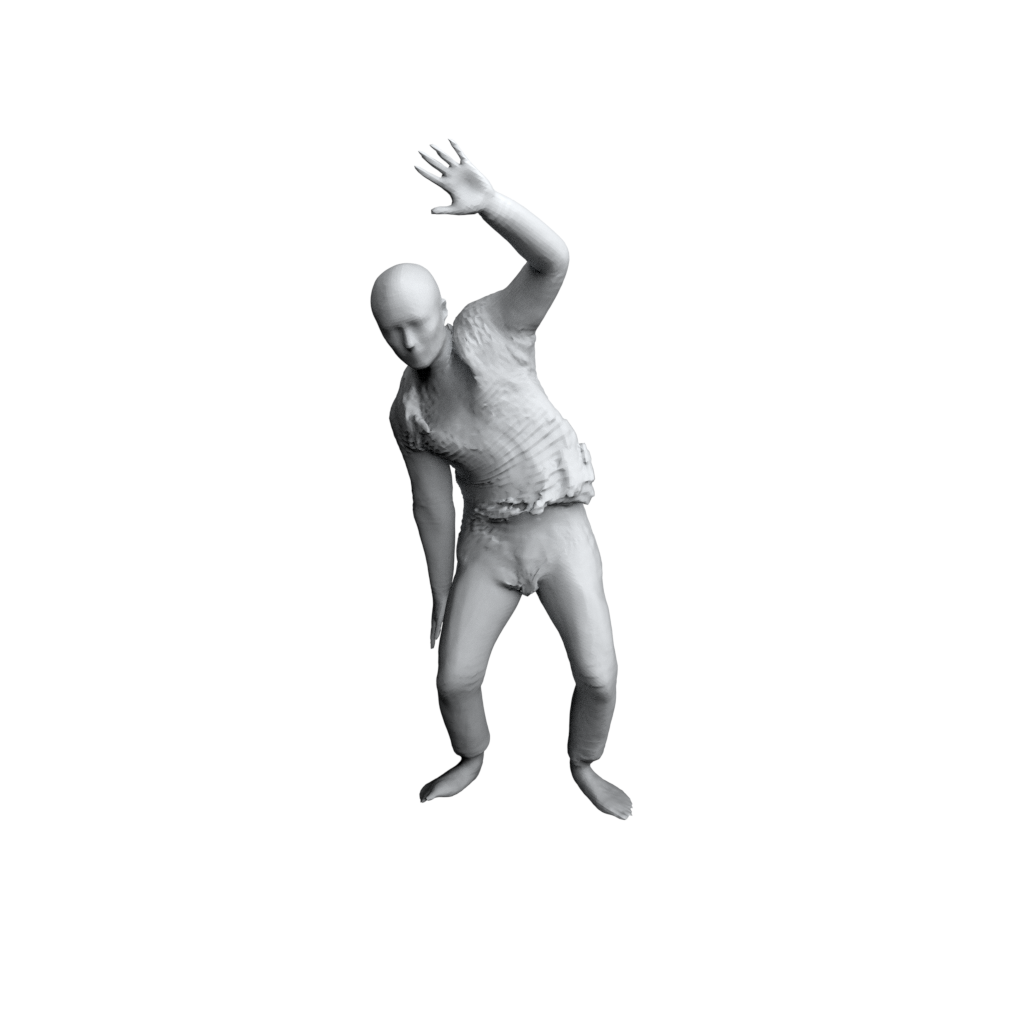}
  \caption{LEAP}
  \end{subfigure}
  \begin{subfigure}{0.3\textwidth}
  \captionsetup{labelformat=empty, font=scriptsize}
    \includegraphics [trim=6cm 7cm 6cm 5cm, width=\textwidth]{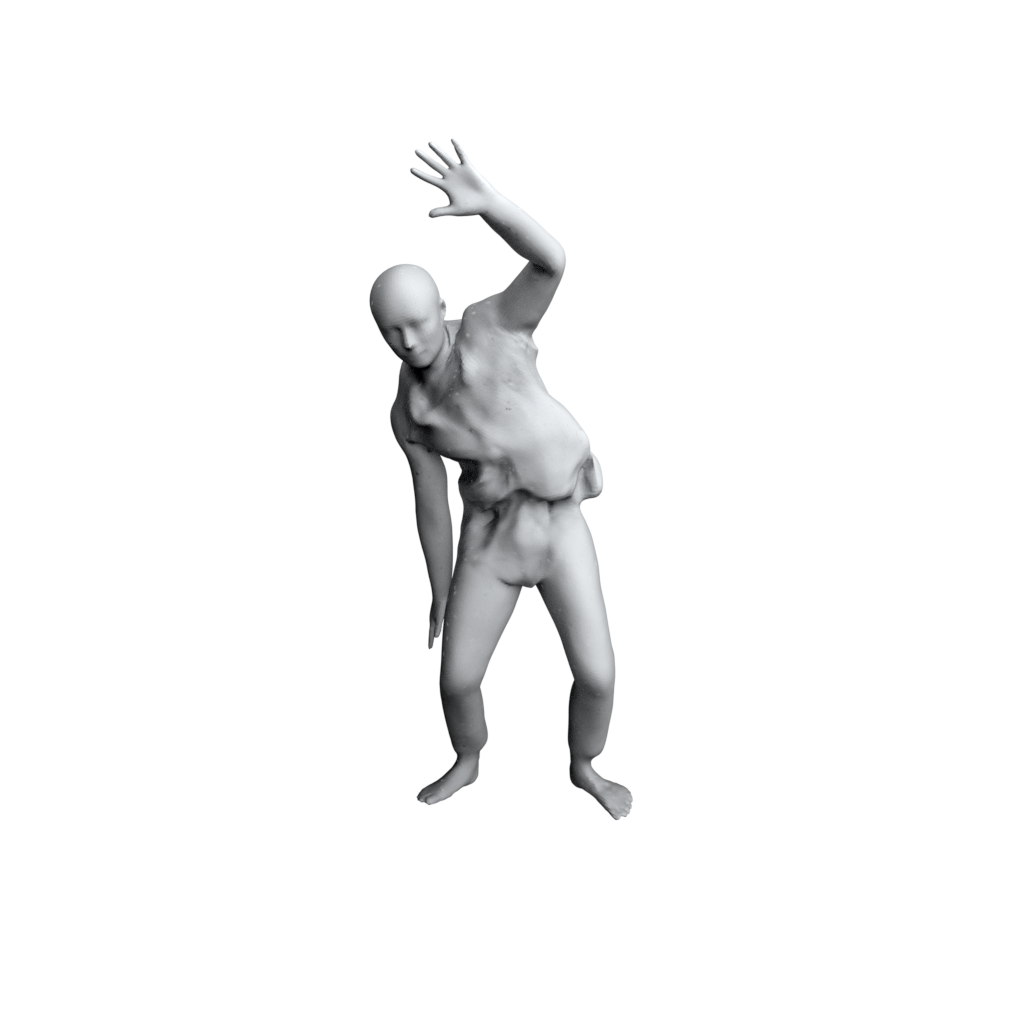}
  \caption{SCANimate}
  \end{subfigure}
  \begin{subfigure}{0.3\textwidth}
  \captionsetup{labelformat=empty, font=scriptsize}
    \includegraphics [trim=6cm 7cm 6cm 5cm, width=\textwidth]{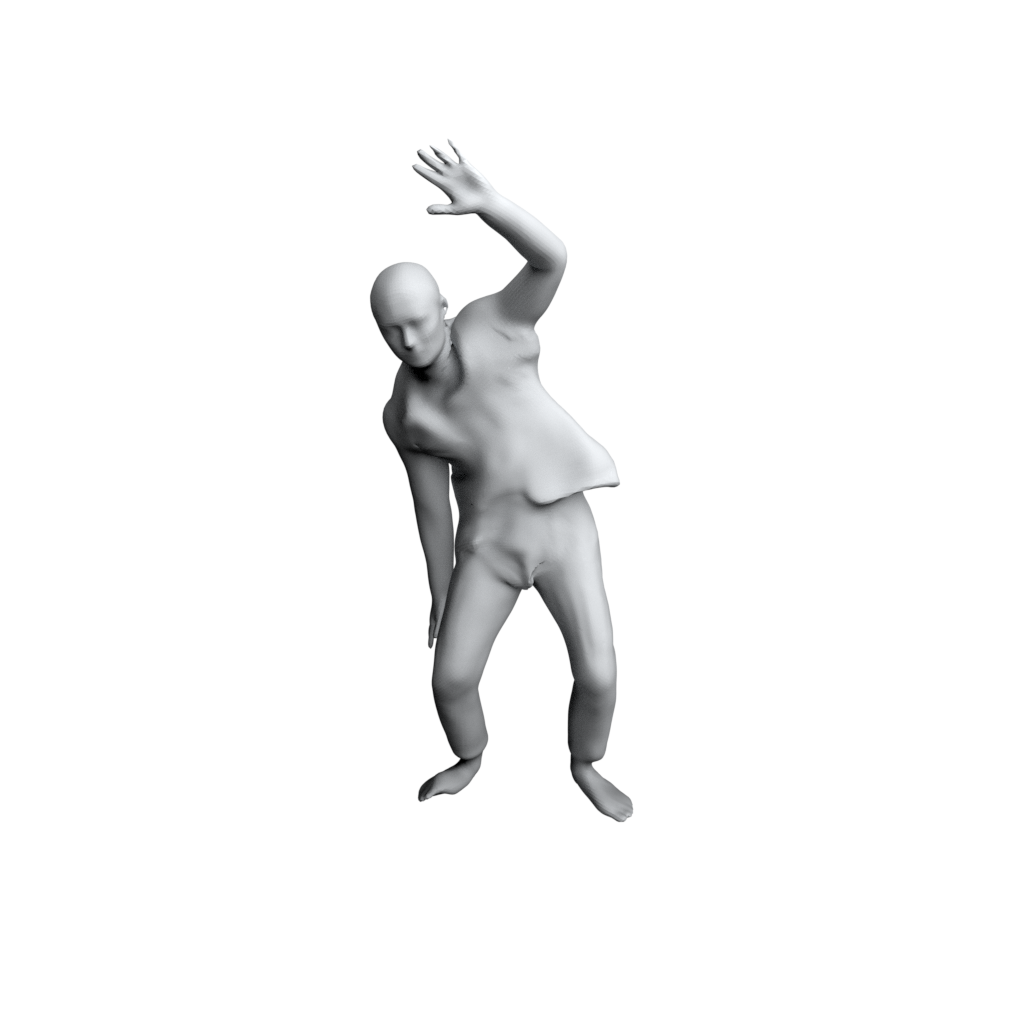}
  \caption{meta-SIREN}
  \end{subfigure}
  \begin{subfigure}{0.3\textwidth}
  \captionsetup{labelformat=empty, font=scriptsize}
    \includegraphics [trim=6cm 7cm 6cm 5cm, width=\textwidth]{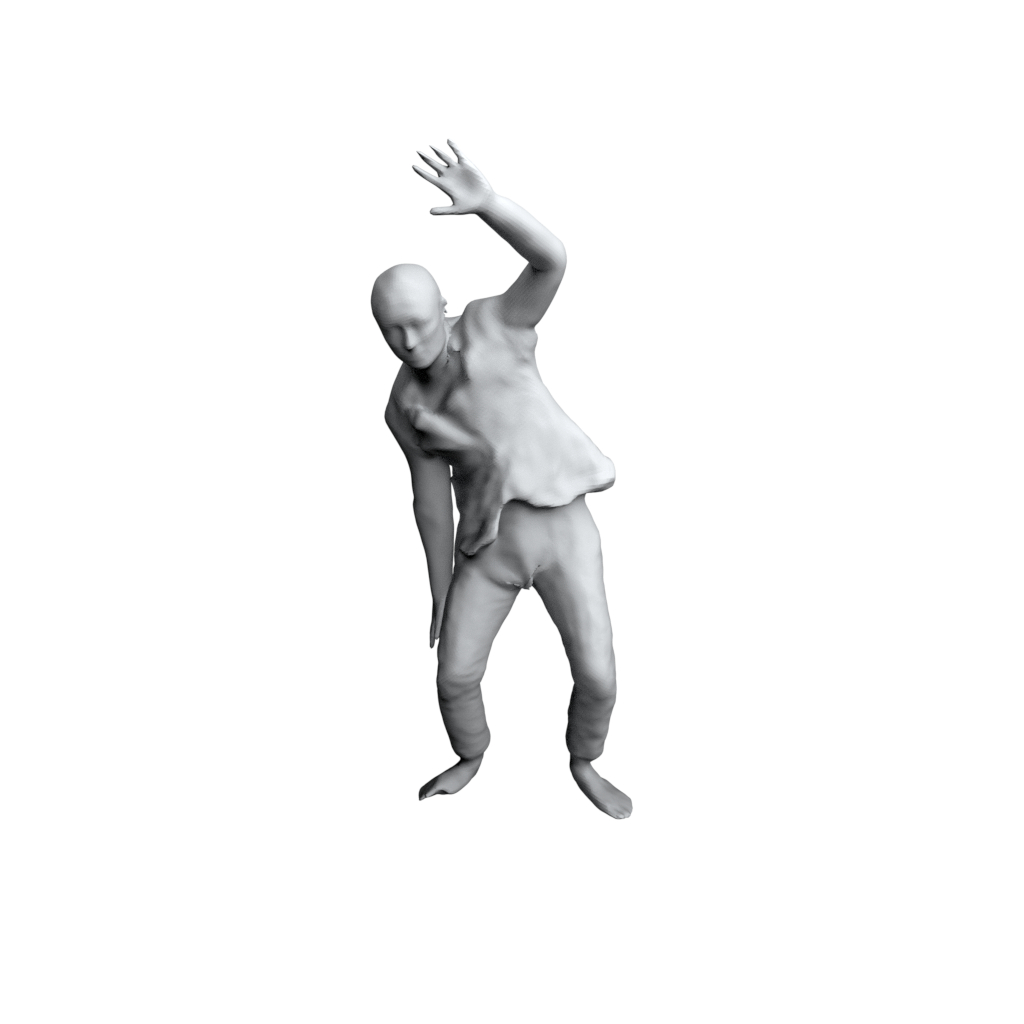}
  \caption{Ours}
  \end{subfigure}
\caption{Qualitative comparison on extrapolation results with blazer outfit. NASA shows consistent blocky artifacts. LEAP predicts overly smooth surfaces missing the tails of the blazer outfit. SCANimate does not generalize as this specific pose has not been seen during training. Directly meta-learning a SIREN~\cite{sitzmann2019siren} network that conditions on input poses produces a smooth surface that does not capture the blazer tails well.}
\label{fig:qualitative_results_baselines}
\end{minipage} \quad
\begin{minipage}[b]{0.48\textwidth}
\centering
  \begin{subfigure}{0.3\textwidth}
  \captionsetup{labelformat=empty, font=scriptsize}
    \includegraphics [trim=5cm 7cm 5cm 5cm, width=\textwidth]{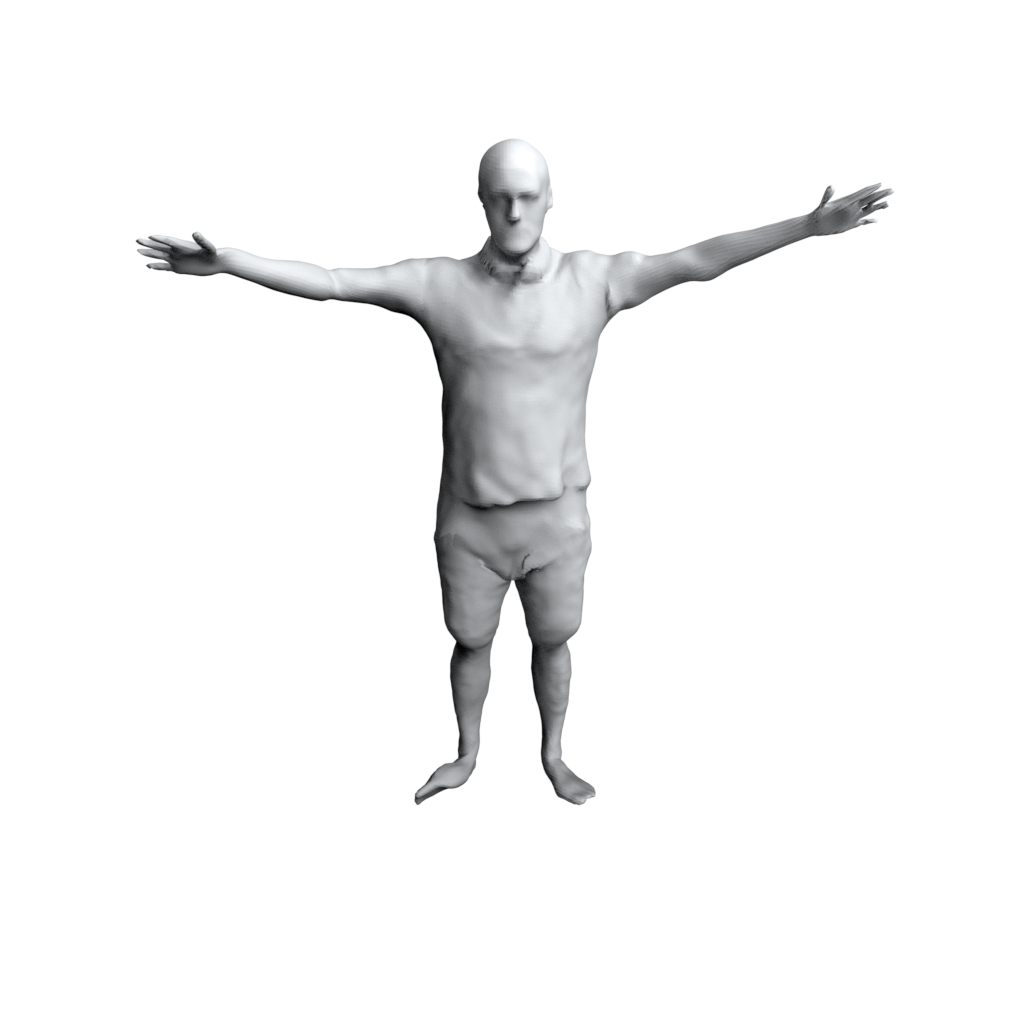}
  \caption{100\%}
  \end{subfigure}
  \begin{subfigure}{0.3\textwidth}
  \captionsetup{labelformat=empty, font=scriptsize}
    \includegraphics [trim=5cm 7cm 5cm 5cm, width=\textwidth]{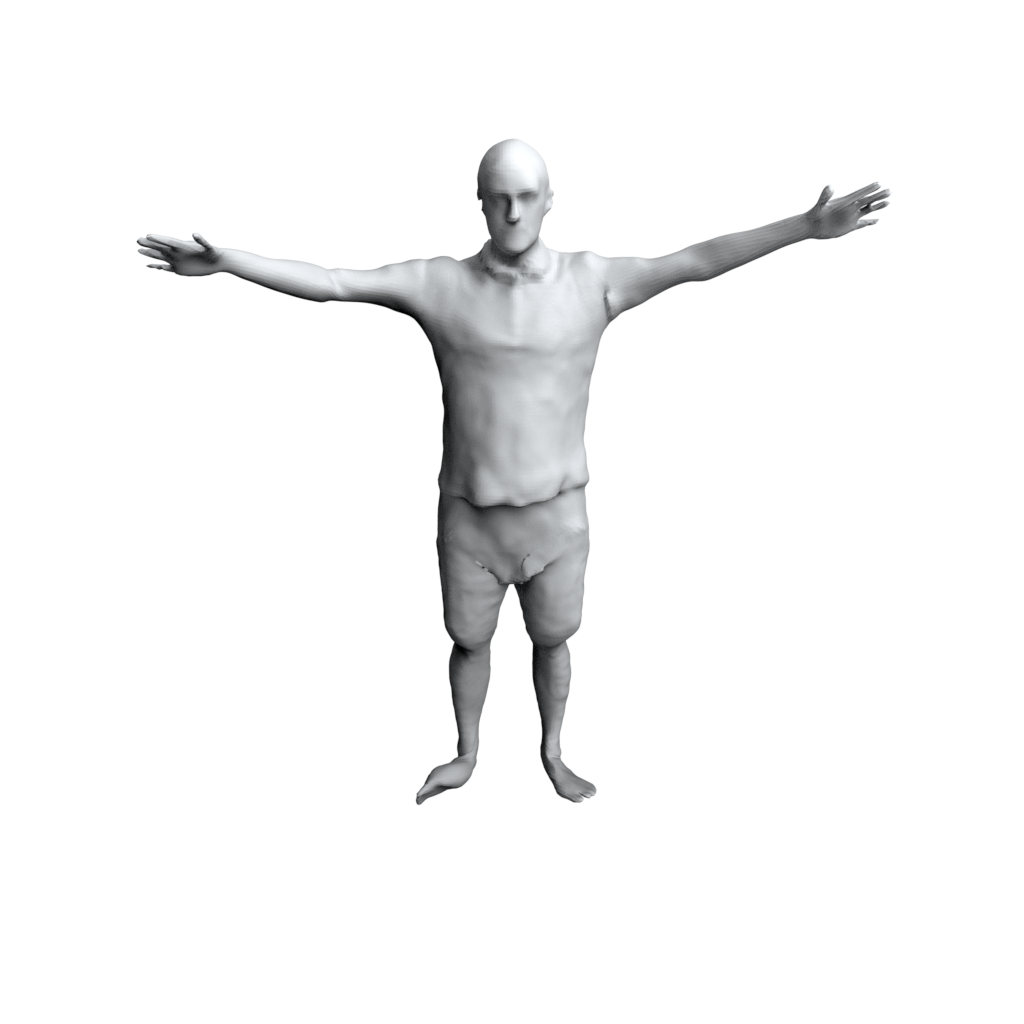}
  \caption{50\%}
  \end{subfigure}
  \begin{subfigure}{0.3\textwidth}
  \captionsetup{labelformat=empty, font=scriptsize}
    \includegraphics [trim=5cm 7cm 5cm 5cm, width=\textwidth]{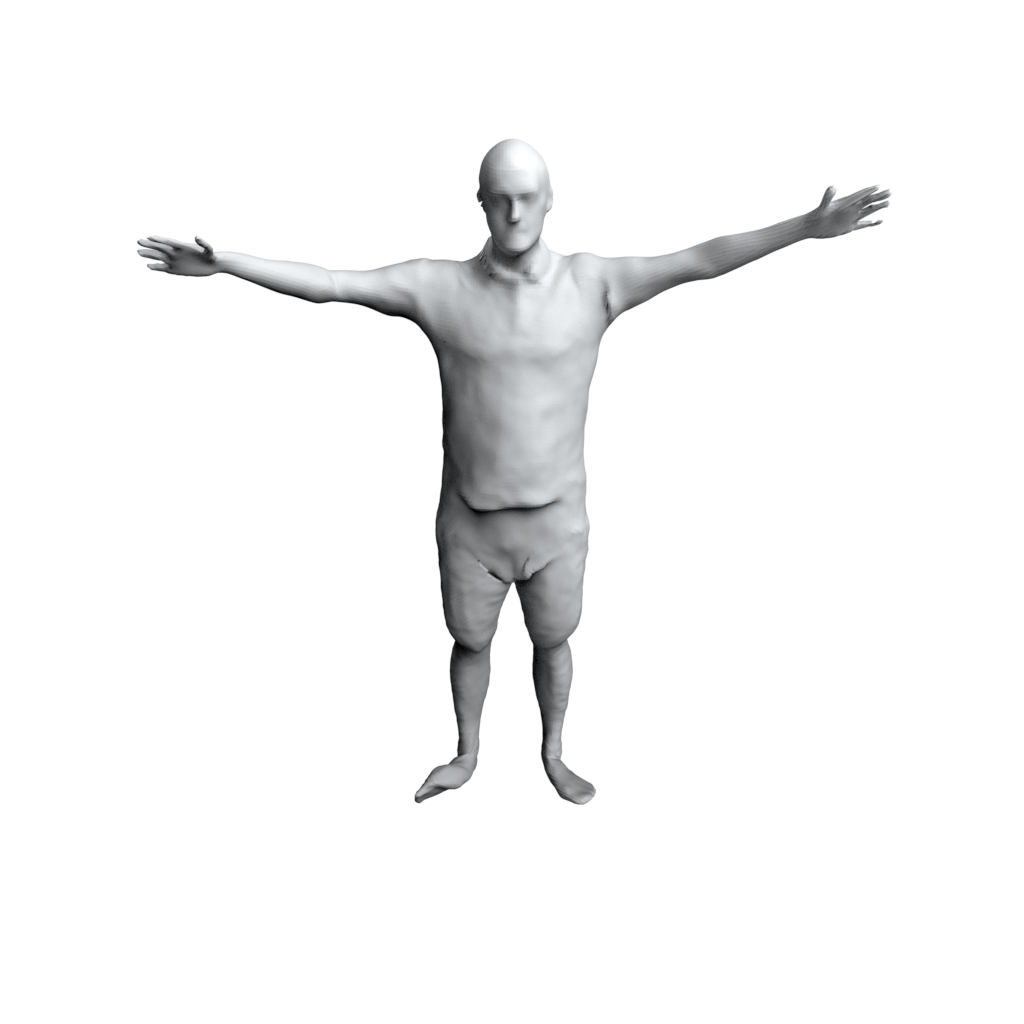}
  \caption{20\%}
  \end{subfigure}
  \begin{subfigure}{0.3\textwidth}
  \captionsetup{labelformat=empty, font=scriptsize}
    \includegraphics [trim=5cm 7cm 5cm 5cm, width=\textwidth]{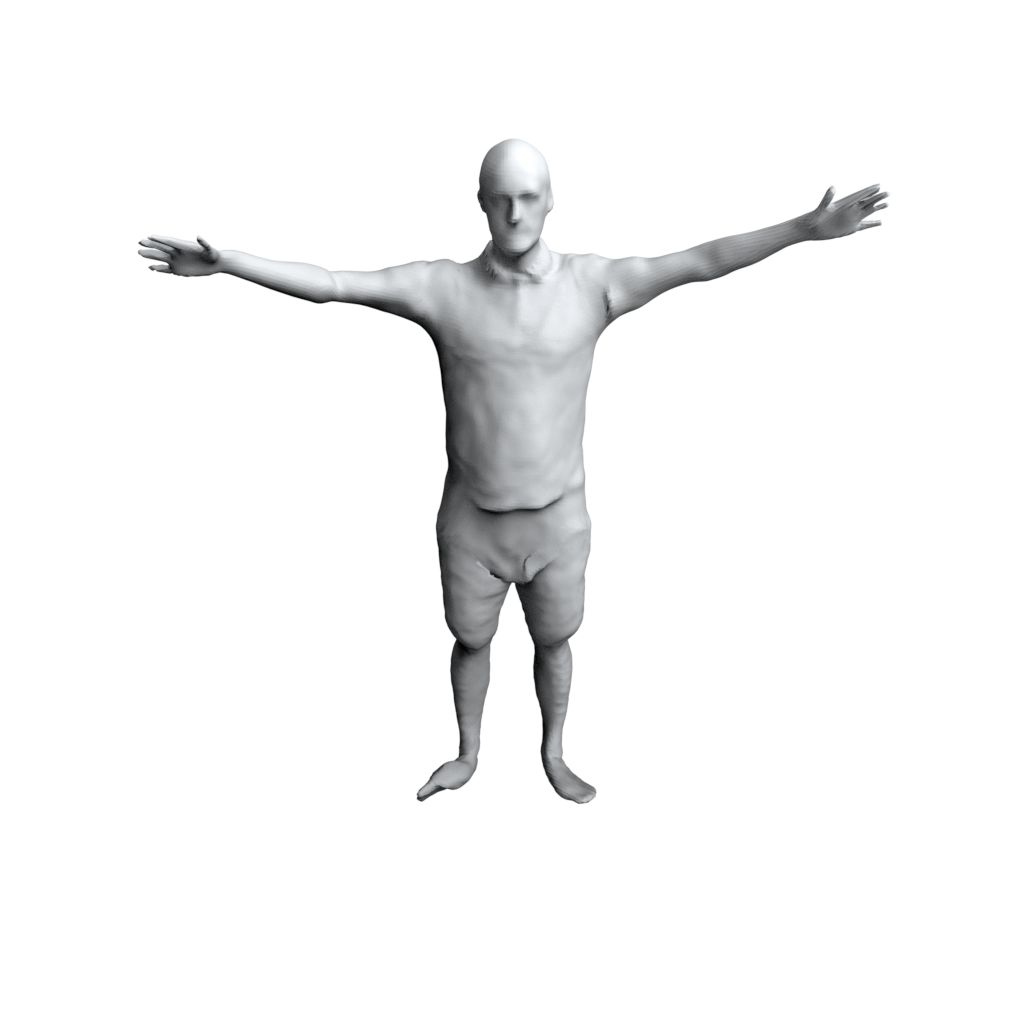}
  \caption{10\%}
  \end{subfigure}
  \begin{subfigure}{0.3\textwidth}
  \captionsetup{labelformat=empty, font=scriptsize}
    \includegraphics [trim=5cm 7cm 5cm 5cm, width=\textwidth]{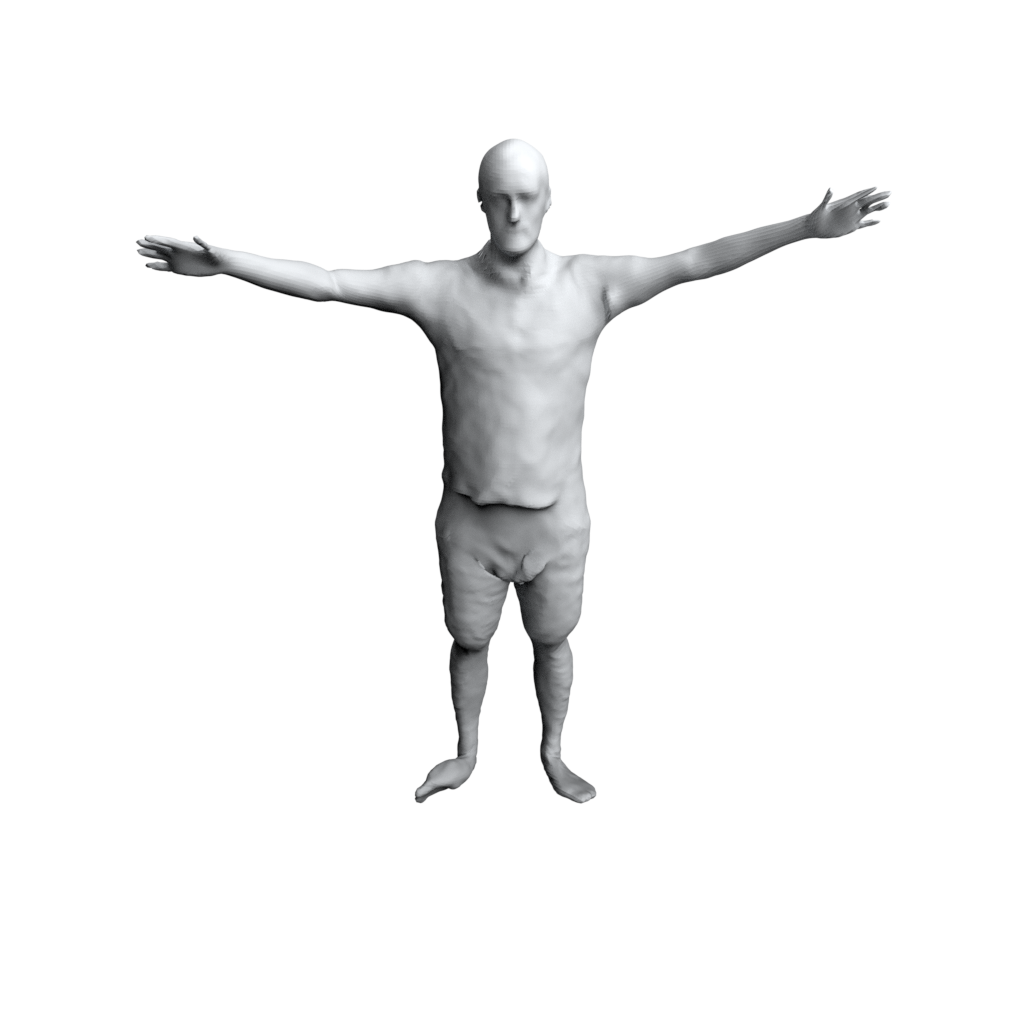}
  \caption{5\%}
  \end{subfigure}
  \begin{subfigure}{0.3\textwidth}
  \captionsetup{labelformat=empty, font=scriptsize}
    \includegraphics [trim=5cm 7cm 5cm 5cm, width=\textwidth]{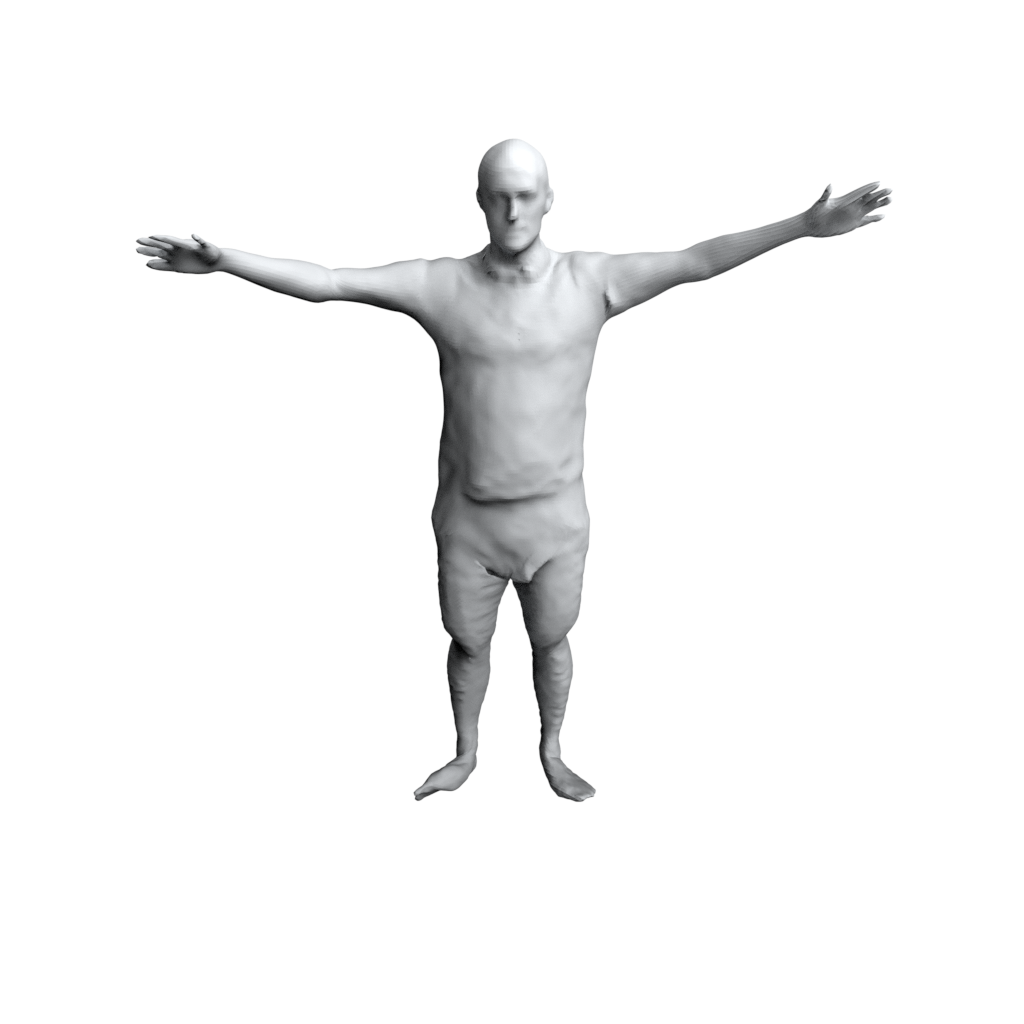}
  \caption{8 depth images}
  \end{subfigure}
\caption{Qualitative comparison on extrapolation results when reducing fine-tuning data on subject 00215 wearing poloshirt. The caption indicates the amount of fine-tuning data used to fine-tune the meta-hypernetwork on this unseen subject. Our meta-learned model captures for this unseen subject the sliding effect of the poloshirt at this pose in which the person raising arms, even fine-tuned with just 8 depth images.}
\label{fig:qualitative_results_fewshot}
\end{minipage}
\end{figure}

Importantly, models of NASA and SCANimate are over-fitted to each subject/cloth-type combination as they cannot straightforwardly leverage prior knowledge from multiple training subjects. 
%SCANimate often fails on poses that were not seen in the fine-tuning set, even with its local pose conditioning approch. 
LEAP is trained on all training and fine-tuning data with input encodings to distinguish different body shapes and cloth types, but it fails to capture high-frequency details of clothes, often predicting smooth surfaces (Fig.~\ref{fig:qualitative_results_baselines}); this is evidenced by its lower perceptual scores (PS) compared to SCANimate and our MetaAvatar. In contrast to these baselines, MetaAvatar successfully captures a unified clothing deformation prior of diverse body shapes and cloth types, which generalizes well to unseen body shapes (00122, 00215), unseen poses (00134, 03375), and unseen cloth types (03375 with blazer outfit); although we did not outperform LEAP on the interpolation task for subject 00134 and 03375, we note that 1) our method uses only 2.5D input for fine-tuning, while LEAP has access to ground-truth canonical meshes during training; 2) subject 00134 and 03375 comprise much more missing frames than subject 00122 and 00215, resulting in higher stochasticity and thus predicting smooth surfaces (such as LEAP) may yield better performance; this is also evidenced by LEAP's much lower perceptual scores on subject 00134 and 03375, although obtaining the best performance for pose interpolation. We encourage the readers to watch the side-by-side comparison videos available on our project page: \href{https://neuralbodies.github.io/metavatar/}{\color{black}{https://neuralbodies.github.io/metavatar/}}.

%\begin{minipage}{.48\linewidth}
%\end{minipage}

\begin{table}
\scriptsize
 \renewcommand{\tabcolsep}{3.0pt}
 \begin{minipage}{.48\linewidth}
 \centering
 \begin{tabular}{ l l c c c c }
 \hline
  \multicolumn{2}{l}{} & \multicolumn{3}{c}{3D Input} & 2.5D Input \\
 \hline
  \multicolumn{2}{l}{} & NASA & LEAP & SCANimate & Ours \\
 \hline
 \multicolumn{6}{c}{Subj 00122, 00215} \\
 \hline
  \ Ex. & \quad PS $\uparrow$ & 0.078 & 0.314 & 0.333 & \textbf{0.5} \\
 \hline
 \multirow{3}{*}{ Int.} & \quad $D_p \downarrow$ & 0.484 & 0.454 & 0.586 & \textbf{0.450} \\
 & \quad $D_f \downarrow$ & 0.327 & 0.293 & 0.489 & \textbf{0.273} \\
 & \quad $NC \uparrow$ & 0.752 & 0.807 & 0.793 & \textbf{0.821} \\
 \hline
 \multicolumn{6}{c}{Subj 00134, 03375} \\
 \hline
 \ Ex. & \quad PS $\uparrow$ & 0.182 & 0.224 & 0.481 & \textbf{0.5} \\
 \hline
 \multirow{3}{*}{ Int.} & \quad $D_p \downarrow$ & 0.595 & \textbf{0.483} & 0.629 & 0.518 \\
 & \quad $D_f \downarrow$ & 0.469 & \textbf{0.340} & 0.542 & 0.367 \\
 & \quad $NC \uparrow$ & 0.693 & \textbf{0.780} & 0.755 & 0.773 \\
 \hline
 \multicolumn{6}{c}{Averge per-model training/fine-tuning time (hours)} \\
 \hline
 & & >10 & - & >10 & 1.60 \\
 \hline
 \end{tabular}
 \vspace{0.05in}
 \caption{\textbf{Comparison to baselines.} $D_p$, $D_f$ and $NC$ are reported for interpolation (Int.) while PS is reported for extrapolation (Ex.). Note that MetaAvatar is fine-tuned on \textit{depth images} while all other baselines are trained on \textit{complete meshes}. The training/fine-tuning times are just rough estimates, as ours does not include the time for meta-learning, while many factors, including varying training schedules, disk-IOs and hardware setups, can affect the final speed.}
 \label{tab:eval_baselines}
 \end{minipage} \quad
 \begin{minipage}{.48\linewidth}
 \centering
 \begin{tabular}{ l l c c c >{\centering\arraybackslash}m{0.3in} >{\centering\arraybackslash}m{0.3in} }
 \hline
  \multicolumn{2}{l}{} & MLP & PosEnc & SIREN & \shortstack{Hyper \\ Quat} & \shortstack{Hyper \\ BoneEnc} \\
 \hline
 \multicolumn{7}{c}{Subj 00122, 00215} \\
 \hline
 \multirow{3}{*}{ Int.} & \quad $D_p \downarrow$ & 3.278 & 1.806 & 0.472 & \textbf{0.460} & 0.461 \\
 & \quad $D_f \downarrow$ & 2.201 & 0.998 & 0.301 & 0.288 & \textbf{0.288} \\
 & \quad $NC \uparrow$ & -0.279 & -0.045 & 0.815 & 0.818 & \textbf{0.820} \\
 \hline
 \multicolumn{7}{c}{Subj 00134, 03375} \\
 \hline
 \multirow{3}{*}{ Int.} & \quad $D_p \downarrow$ & 3.320 & 1.498 & 0.532 & 0.526 & \textbf{0.523} \\
 & \quad $D_f \downarrow$ & 2.190 & 0.772 & 0.385 & 0.378 & \textbf{0.374} \\
 & \quad $NC \uparrow$ & -0.300 & -0.099 & \textbf{0.773} & 0.772 & 0.772 \\
 \hline
 \end{tabular}
 % }
 \vspace{0.05in}
 \caption{\textbf{Ablation for different architectures on the interpolation task.} Hyper-Quat is our model that takes the relative joint-rotations (in the form of unit quaternions) as inputs. Hyper-BoneEnc is our full model with hierarchical bone encoding MLP of LEAP~\cite{LEAP:CVPR:21}. Models in the table are fine-tuned for 128 epochs.}
 \label{tab:eval_arch}
 \end{minipage}
\end{table}

\subsection{Ablation Study on Model Architectures}
\label{sec:exp_arch}
We further ablate model architecture choices for MetaAvatar. We compare against (1) a plain MLP that takes the concatenation of the relative joint-rotations (in the form of unit quaternions) and query points as input (MLP), (2) a MLP that takes the concatenation of the relative joint-rotations and the positional encodings of query point coordinates as input (PosEnc), and (3) a SIREN network that takes the concatenation of  the relative joint-rotations and query points as input (SIREN). 
The evaluation task is interpolation; results are reported in \tablename~\ref{tab:eval_arch}. For the baselines (MLP, PosEnc and SIREN), we directly use Alg.~\ref{alg:reptile_hyper-sdf} to meta-learn the corresponding models with $\phi^*=0$. For MLP and PosEnc, the corresponding models fail to produce reasonable shapes. For SIREN, it produces unnaturally smooth surfaces which cannot capture fine clothing details such as wrinkles (Fig.~\ref{fig:qualitative_results_baselines}).

\subsection{Few-shot learning of MetaAvatar}
\label{sec:exp_fewshot}
\begin{wraptable}[16]{R}{6.7cm}
    \scriptsize
     \vspace{-10pt}
    \setlength{\tabcolsep}{3pt}
 \centering
 \begin{tabular}{ l l c c c c c c }
 \hline
  \multicolumn{2}{l}{Fine-tune data (\%)} & 100 & 50 & 20 & 10 & 5 & <1 \\
 \hline
 \multicolumn{8}{c}{Subj 00122, 00215} \\
 \hline
  \ Ex. & \quad PS $\uparrow$ & 0.5 & 0.471 & 0.509 & 0.473 & 0.373 & \textbf{0.510} \\
 \hline
 \multirow{3}{*}{ Int.} & \quad $D_p \downarrow$ & - & \textbf{0.450} & 0.480 & 0.512 & 0.543 & 0.592 \\
 & \quad $D_f \downarrow$ & - & \textbf{0.273} & 0.310 & 0.353 & 0.391 & 0.450 \\
 & \quad $NC \uparrow$ & - & \textbf{0.821} & 0.808 & 0.795 & 0.785 & 0.768 \\
 \hline
 \multicolumn{8}{c}{Subj 00134, 03375} \\
 \hline
 \ Ex. & \quad PS $\uparrow$ & \textbf{0.5} & 0.476 & 0.424 & 0.463 & 0.439 & 0.387 \\
 \hline
 \multirow{3}{*}{ Int.} & \quad $D_p \downarrow$ & - & \textbf{0.518} & 0.545 & 0.576 & 0.603 & 0.619 \\
 & \quad $D_f \downarrow$ & - & \textbf{0.367} & 0.400 & 0.438 & 0.471 & 0.489 \\
 & \quad $NC \uparrow$ & - & \textbf{0.773} & 0.762 & 0.753 & 0.745 & 0.737 \\
 \hline
 \multicolumn{8}{c}{Average per-model training/fine-tuning time (hours)} \\
 \hline
 & & 1.60 &  0.8 & 0.32 & 0.16 & 0.08 & 0.02 \\
 \hline
 \end{tabular}
%  \vspace{1em}
 \caption{\textbf{Ablation for few-shot learning.} We report performance of MetaAvatar on reduced amount of fine-tuning data. Fine-tuning time scales linearly with the amount of data, since we run for a fixed number of epochs.} %Note that for interpolation (Int.) experiments in Section~\ref{sec:exp_baselines} we train on every 10th frame, which is equivalent to 50\% of the data used to train full extrapolation (Ex.) models.}
 \label{tab:eval_fewshot}
\end{wraptable}
In this section, we evaluate the few-shot learning capabilities of MetaAvatar. As shown in \tablename~\ref{tab:eval_fewshot}, we reduce the amount of data on the fine-tuning set, and report the performance of models fine-tuned on reduced amount of data. Note that with $<\!1\%$ data, we require only one frame from each action sequence available for a subject/cloth-type combination, this roughly equals to 8-20 depth frames depending on the amount of data for that subject/cloth-type combination. For interpolation, the performance drops because the stochastic nature of cloth deformation becomes dominant when the amount of fine-tuning data decreases. On the other hand, the perceptual scores (PS) are better than NASA and LEAP even with \textbf{<1\%} data in the form of \textbf{partial depth observations}, and better than or comparable to SCANimate in most cases. The qualitative comparison on extrapolation results of reduced fine-tuning data is shown in Fig.~\ref{fig:qualitative_results_fewshot}. Please see Appendix~\ref{appx:additional_quatlitative} for more qualitative results on few-shot learning, including results on depth from raw scans, results on real depth images and comparison with pre-trained SCANimate model.

\section{Conclusion}
\label{sec:conclusion}
We introduced MetaAvatar, a meta-learned hypernetwork that represents controllable dynamic neural SDFs applicable for generating clothed human avatars. Compared to existing methods, MetaAvatar learns from less data (temporally discrete monocular depth frames) and requires less time to represent novel unseen clothed humans. 
%, all thanks to its meta-learned shape priors. 
We demonstrated that the meta-learned deformation prior is robust and can be used to effectively generate realistic clothed human avatars in 2 minutes from as few as 8 depth observations. 

% impact (future work) & broader impact
% In the future, we plan to combine automatic registration methods~\cite{bhatnagar2020loopreg,PTF:CVPR:2021} to create an end-to-end learning pipeline for clothed human reconstruction and avatar creation from in-the-wild RGBD sequences.
MetaAvatar is compatible with automatic registration methods~\cite{bhatnagar2020loopreg,PTF:CVPR:2021}, human motion models~\cite{Zhang:CVPR:2021,Zhang:ICCV:2021}, implicit hand models~\cite{karunratanakul2020grasping,karunratanakul2021skeletondriven} and rendering primitives~\cite{MVP:TOG:21, DeepSurfels:CVPR:21} that could jointly enable an efficient end-to-end photo-realistic digitization of humans from commodity RGBD sensors, which has broad applicability in movies, games, and telepresence applications. 
However, this digitization may raise privacy concerns that need to be addressed carefully before deploying the introduced technology.

\section{Acknowledgment}
\label{sec:acknowledgement}
Siyu Tang acknowledges funding by the Swiss National Science Foundation under project 200021\_204840. Andreas Geiger was supported by the ERC Starting Grant LEGO-3D (850533) and DFG EXC number 2064/1 - project number 390727645. We thank Jinlong Yang for providing results of SCANimate on the CAPE dataset. We thank Yebin Liu for sharing POSEFusion~\cite{li2021posefusion} data and results during the rebuttal period. We also thank Yan Zhang, Siwei Zhang and Korrawe Karunratanakul for proof reading the paper.

% \clearpage
{\small
\bibliographystyle{ieee_fullname}
\bibliography{egbib}
}

%%%%%%%%%%%%%%%%%%%%%%%%%%%%%%%%%%%%%%%%%%%%%%%%%%%%%%%%%%%%
\clearpage

\appendix
\numberwithin{equation}{section}
\setcounter{equation}{0}
\numberwithin{figure}{section}
\setcounter{figure}{0}
\numberwithin{table}{section}
\setcounter{table}{0}
\numberwithin{algorithm}{section}
\setcounter{algorithm}{0}

\section{Loss Definitions and Hyperparameters for Skinning Networks}
\label{appx:skinning_nets}
In 
Section~\ref{sec:lbs_net},
%Section~\textcolor{red}{3} of the main paper,
we followed~\cite{LEAP:CVPR:21,SCANimate:CVPR:21} and defined the loss for training the skinning networks as follows:
\begin{align}
\mathcal{L} (\mathbf{X}) = \lambda_{r} \mathcal{L}_{r} + \lambda_{s} \mathcal{L}_{s} + \lambda_{skin} \mathcal{L}_{skin} \,,
\end{align}
where $\mathcal{L}_r$ represents a re-projection loss that penalizes the $l_2$ distance between an input point $\mathbf{x}$ and the corresponding re-projected point $\bar{\mathbf{x}}$:
\begin{align}
\label{eqn:reproj_loss}
\mathcal{L}_r (\mathbf{X}) &= \frac{1}{N} \sum_{i=1}^{N} \lVert \mathbf{x}^{(i)} - \bar{\mathbf{x}}^{(i)}  \rVert_{2}\,,
\end{align}
$\mathcal{L}_s$ is the mean $l_1$ distances between the predicted forward skinning weights and inverse skinning weights:
\begin{align}
\mathcal{L}_s (\mathbf{X}) &= \frac{1}{N} \sum_{i=1}^{N} \sum_{b=1}^{24} | h_{\text{fwd}}(\hat{\mathbf{X}}, \hat{\mathbf{x}}^{(i)})_{b} - h_{\text{inv}}(\mathbf{X}, \mathbf{x}^{(i)})_{b} |\,,
\end{align}
and $\mathcal{L}_{skin}$ represents the mean $l_1$ distance between the predicted (forward and inverse) skinning weights and the barycentrically interpolated skinning weights $\mathbf{w}^{(i)}$ on the registered minimally-clothed shape that is closest to point $\mathbf{x}^{(i)}$:
\begin{align}
\mathcal{L}_{skin} (\mathbf{X}) = \frac{1}{N} \sum_{i=1}^{N} \sum_{b=1}^{24} | h_{\text{inv}}(\mathbf{X}, \mathbf{x}^{(i)})_{b} - \mathbf{w}^{(i)}_b | \nonumber + | h_{\text{fwd}}(\hat{\mathbf{X}}, \hat{\mathbf{x}}^{(i)})_{b} - \mathbf{w}^{(i)}_b | \,.
\end{align}
We empirically set $\lambda_r = \lambda_s = 1$ and $\lambda_{skin} = 10$ throughout our experiments. For all models (single-view point cloud and full point cloud), we use the Adam~\cite{Adam:ICLR:2015} optimizer with a learning rate of $1e^{-4}$ and train on the training set (10 subjects of CAPE) for 200k iterations, with a mini-batch size of 12. The learning rate of $1e^{-4}$ is a common choice used in many previous works~\cite{Occupancy_Networks,sitzmann2019metasdf,ConvONet} and we did not specifically tune it. 

\section{Hyperparameters for MetaAvatar}
\label{appx:metavatar}
In this section, we describe the hyperparameters used for our MetaAvatar model. We specify the hyperparameters for the two algorithms described in Section~\ref{sec:meta-sdf} and Section~\ref{sec:meta-hyper-sdf}, respectively.

\boldparagraph{Meta-learned Initialization of Static Neural SDFs (Sec.~\ref{sec:meta-sdf})} The algorithm is described as Alg.~\ref{appx:alg:reptile_sdf}:
 
\begin{algorithm}
%\scriptsize
\caption{Meta-learning SDF with Reptile~\cite{Reptile:arXiv:2018}}
\begin{algorithmic}[1] 
  \Statex \textbf{Initialize}: meta-network parameters $\phi$, meta learning rate $\beta$, inner learning rate $\alpha$, max training iteration $N$, inner-loop iteration $m$, batch size $M$
  \For{$i = 1, \ldots, N$}
    \State Sample a batch of $M$ training samples $\{ \hat{\mathbf{X}}^{(j)} \}_{j=1}^{M}$
    \For {$j = 1, \ldots, M$}
      \State $\phi^{(j)}_0 = \phi$
      \For{$k = 1, \ldots, m$}
        \State $\phi^{(j)}_{k} = \phi^{(j)}_{k-1} - \alpha \nabla_{\phi}\mathcal{L}_{\text{IGR}}(f_{\phi}(\hat{\mathbf{X}}^{(j)})|_{\phi=\phi^{(j)}_{k-1}})$ \label{appx:alg:meta-sdf_inner_update}
      \EndFor
    \EndFor
    \State $\phi \leftarrow \phi + \beta \frac{1}{M} \sum_{j=1}^{M} (\phi^{(j)}_m - \phi)$ \label{appx:alg:meta-sdf_outer_update}
  \EndFor
\end{algorithmic}
\label{appx:alg:reptile_sdf}
\end{algorithm}

where we set $\beta = 1e^{-5}$, $\alpha = 1e^{-4}$, $N = 160k$, $m = 24$ and $M = 3$. $m = 24$ is empirically set. The learning rates are found by starting at $1e^{-4}$ then decrease it by 10 times each time the model cannot converge. Same applies for hyperparameters of Alg.~\ref{appx:alg:reptile_hyper-sdf} and other ablation baselines. For both of the outer-loop update (line~\ref{appx:alg:meta-sdf_outer_update}) and inner-loop update (line~\ref{appx:alg:meta-sdf_inner_update}) we use the Adam~\cite{Adam:ICLR:2015} optimizer as we found that the SGD optimizer does not converge, probably due to the complex second-order gradients caused by the IGR~\cite{Gropp:2020:ICML} loss.

\boldparagraph{Meta-learned Initialization of HyperNetwork for Dynamic Neural SDFs (Sec.~\ref{sec:meta-hyper-sdf})} The algorithm is described as Alg.~\ref{appx:alg:reptile_hyper-sdf}.
We set $\beta = 1e^{-6}$, $\alpha = 1e^{-6}$, and $m = 24$. Since the number of samples per inner-loop is random, we did not specify the maximum number of training iterations $N$, but rather train for 50 epochs. Again we use Adam for gradient updates of both the inner and the outer loop. For each MLP of the hypernetwork, we initialize the last linear layer with zero weights and biases, while all other linear layers are initialized using the default setting of PyTorch~\cite{pytorch}, \ie\ the He initialization~\cite{HeInit:ICCV:2015}.
 
\begin{algorithm*}
%\scriptsize
\caption{Meta-learning hypernetwork with Modified Reptile}
\begin{algorithmic}[1] 
  \Statex \textbf{Initialize}: meta-hypernetwork parameters $\psi$, pre-trained meta-SDF parameters $\phi^*$, meta learning rate $\beta$, inner learning rate $\alpha$, max training iteration $N$, inner-loop iteration $m$.
  \For{$i = 1, \ldots, N$}
    \State $\psi_0 = \psi$
    \State Randomly choose a subject/cloth-type combination $n$
    \State Uniformly sample $M \sim \{1, \ldots, D^{(n)}\}$ where $D^{(n)}$ is the number of datapoints of subject/cloth-type combination $n$
    \State Sample $M$ datapoints from subject/cloth-type combination $n$, denoting these datapoints as $\mathcal{S} = \{ \{ \mathbf{B}_b \}^{(j)}, \hat{\mathbf{X}}^{(j)} \}_{j=1}^{M}$ 
    \For{$k = 1, \ldots, m$}
      \State $\mathcal{L} = \frac{1}{M} \sum_{(\{ \mathbf{B}_b \}, \hat{\mathbf{X}}) \in \mathcal{S}} \mathcal{L}_{\text{IGR}}(f_{\phi}(\hat{\mathbf{X}})|_{\phi=g_{\psi_{k-1}}(\{ \mathbf{B}_b \}) + \phi^*})$ \label{appx:alg:hyper-meta-loss}
      \State $\psi_{k} = \psi_{k-1} - \alpha \nabla_{\psi_{k-1}} \mathcal{L}$ \label{appx:alg:hyper-meta-inner}
    \EndFor
    \State $\psi \leftarrow \psi + \beta (\psi_{m} - \psi_0)$
  \EndFor
\end{algorithmic}
\label{appx:alg:reptile_hyper-sdf}
\end{algorithm*}

\boldparagraph{Implementation Details} For the IGR loss we use the same hyperparameters as in SIREN~\cite{sitzmann2019siren} for both algorithms. We normalize all query points in the canonical space to the unit cube $[-1, 1]^3 $ according to the underlying SMPL registration of each frame. For depth rendering, we render the meshes into $250 \times 250$ depth images, which end up having roughly 3000-5000 points per frame; for both training and fine-tuning, we randomly duplicate (if the frame has less than 5000 points) or sample (if the frame has more than 5000 points) these points to make each frame contains exactly 5000 points such that we can process them in batches. For off-surface points, we randomly sample 5000 points in the unit cube $[-1, 1]^3$. 

During fine-tuning, we remove all points whose re-projection distance (Eq.~\eqref{eqn:reproj_loss}) is greater than 2cm.

Very occasionally (a dozen of frames out of more than 5000 frames), we observe visual artifacts caused by floating blobs which are the results of imperfect SDFs. We remove such floating blobs by keeping only the largest connected component of the extracted iso-surfaces. This post-processing does not impact the mesh-distance and normal consistency metrics as they consist only a very small portion of the results.
\section{Hyperparameters for Baselines}
\label{appx:baselines}
We further describe the hyperparameters for baselines that we have compared to in Section~\ref{sec:exp_baselines}.
%~\textcolor{red}{5} of the main paper.

\subsection{NASA}
We follow NASA~\cite{Deng_ECCV2020} and train each model for 200k iterations while using the Adam optimizer~\cite{Adam:ICLR:2015} with the learning rate of $1e^{-4}$ and the batch size of 12. Each batch consists of 1024 uniformly sampled points and 1024 points sampled around the mesh surface; an additional set of 2048 points are sampled on the ground-truth mesh surface to compute the weight auxiliary loss.

\subsection{LEAP}
We follow the experimental setup presented in LEAP~\cite{LEAP:CVPR:21} and train it until convergence ($\approx$ 300k iterations) by using the Adam optimizer~\cite{Adam:ICLR:2015} (learning rate $1e^{-4}$) with the batch size of 30. 
Each batch consists of 2048 uniformly sampled points and 2048 points sampled around the mesh surface in the canonical space.

Since LEAP is designed for minimally-clothed human bodies, we extend it to learn more challenging clothing types. 
Specifically, we represent different clothing types via a one-hot encoding vector $\mathbb{R}^9$ and concatenate it to the LEAP's global feature vector.  

\subsection{SCANimate}
We follow the architecture and hyperparameter settings as provided in the SCANimate~\cite{SCANimate:CVPR:21} paper, and train the models in three stages: 80 epochs for pre-training the skinning nets, 200 epochs for training the skinning nets with the cycle-consistency loss, and 1000 epochs for training the neural SDF. For the ground truth clothed body data, we sample 8000 points on the clothed body meshes, and do so dynamically at each iteration.

Note that the full SCANimate pipeline is designed to work on raw scans, whereas in our settings, the clothed body meshes are registered meshes with the SMPL model topology (6890 vertices and 13776 triangles). 
We strictly follow the full pipeline of SCANimate, which includes a step of removing distorted triangles (up to an edge stretching threshold of 2.0) after canonicalization. 
While this step works well for high-resolution raw scans, it may unnecessarily remove certain pose-dependent deformations 
when applying to the registered clothed body meshes in our settings.
This happens most frequently around the armpit and elbow region, which, consequently, creates artifacts at these regions in the final posed body meshes. 
Such domain gap in the data can help explain the slightly compromised quantitative performance of SCANimate in Section~\ref{sec:exp_baselines}.
% ~\textcolor{red}{5.1} of the main paper.
Nonetheless, SCANimate shows impressive qualitative results and perceptual scores, and yet our model still outperforms it in both metrics, demonstrating the effectiveness of our approach.
\section{Hyperparameters for Ablation Baselines}
\label{appx:ablation}
In this section, we describe the hyperparameters for ablation baselines that we have compared to in Section~\ref{sec:exp_arch}. To recap, the ablation baselines include (1) a plain MLP that takes the concatenation of the relative joint-rotations (in the form of unit quaternions) and query points as input (MLP), (2) an MLP that takes the concatenation of the relative joint-rotations and the positional encodings of query point coordinates as input (PosEnc), and (3) a SIREN network that takes the concatenation of the relative joint-rotations and query points as input (SIREN). 

\boldparagraph{Shared Hyperparameters} All ablation baselines (MLP, PosEnc and SIREN) are trained using Alg.~\ref{appx:alg:reptile_hyper-sdf} with $\phi^* = 0$. We set $\beta = 1e^{-5}$, $\alpha = 1e^{-4}$ and $m = 24$. Since these ablation baselines do not have an initialization like MetaAvatar does, we train them for 100 epochs instead of 50 epochs.

\boldparagraph{PosEnc} We use the 8th-order positional encoding~\cite{mildenhall2020nerf} to encode the xyz-coordinates while leaving the pose inputs as it is.

\boldparagraph{SIREN} We initialize the weights of the network the same way as described in the original SIREN~\cite{sitzmann2019siren} paper.

\section{Perceptual Study Details}
\label{appx:perceptual_study}
\begin{figure}
    \centering
    \includegraphics[width=\textwidth]{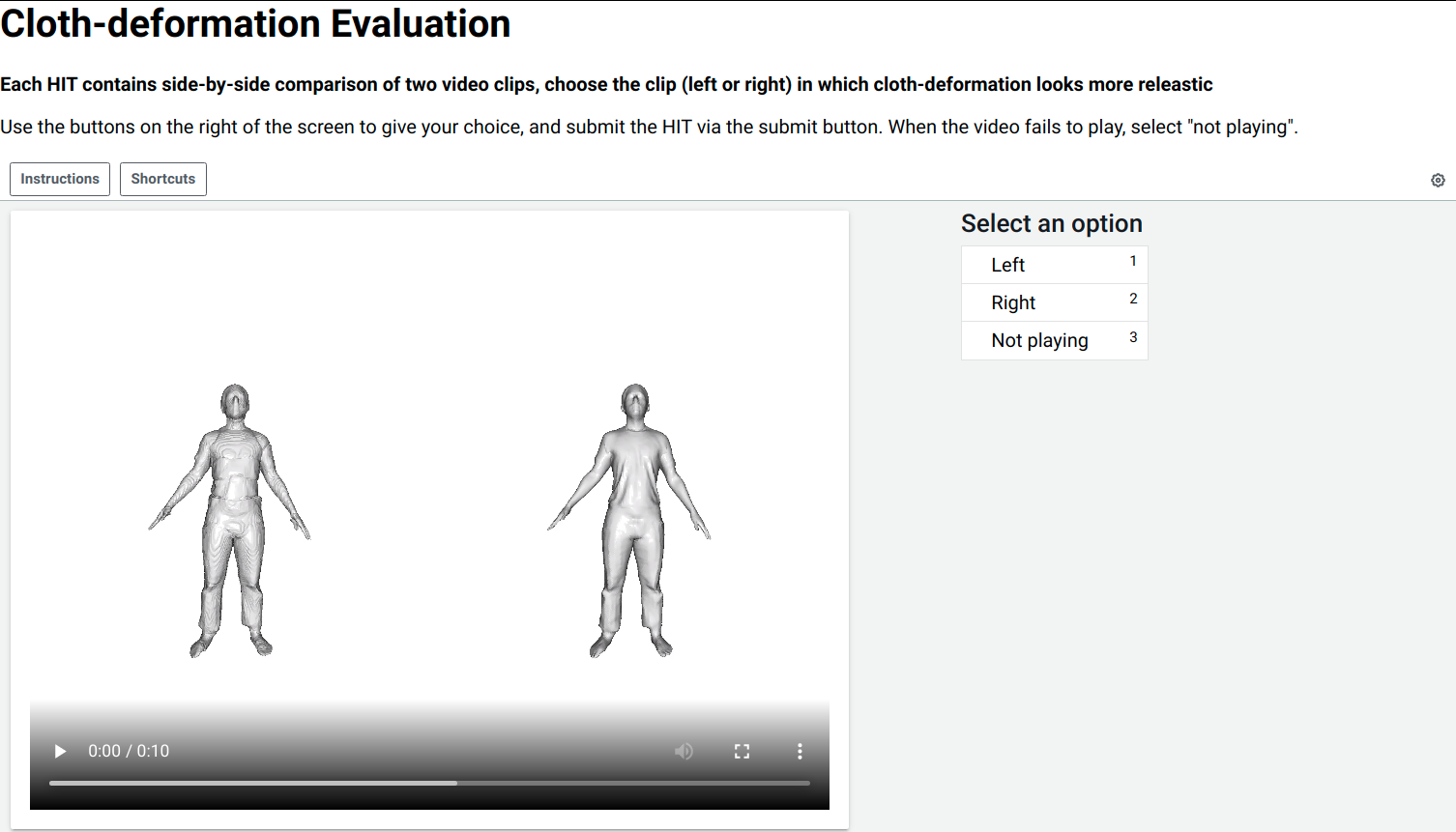}
    \caption{User interface of our perceptual study}
    \label{fig:UI_perceptual_study}
\end{figure}

The Amazon Mechanical Turk (AMT) user interface and instruction of our perceptual study are illustrated in Fig.~\ref{fig:UI_perceptual_study}. In each Human Intelligence Task (HIT), the user is provided with a side-by-side comparison video of MetaAvatar versus one baseline approach, the video corresponds to one action sequence of CAPE~\cite{CAPE:CVPR:20} which lasts 5-10 seconds. The left-right ordering is randomly shuffled for each video so that users do not become biased towards one side. To take internet/OS-related issues into account, we also provide a "Not playing" option in case certain users cannot play some videos on their devices. For each video, we ask 3 different users to independently choose the side on which they think the cloth-deformation looks more realistic. Given the total number of videos as $M$, the perceptual score is calculated as $PS = \frac{P}{3M - N}$, where $N$ is the number of user choices that choose "Not playing", and $P$ is the number of user choices that choose the baseline approach over MetaAvatar. Thus the perceptual score is in the $[0, 1]$ range, a score that is less than $0.5$ means the baseline looks less realistic than MetaAvatar, while a score that is greater than $0.5$ indicates that the baseline looks more realistic than MetaAvatar. For MetaAvatar, we manually define its perceptual score to be $0.5$.

For salary, we provide a 0.02\$ reward for each HIT. For evaluation against baselines, we evaluated a total number of 1143 HITs, which costs 34.29\$ (22.86\$ for reward and 11.43\$ for AMT). For evaluation of few-shot learning, we evaluated a total number of 1905 HITs, which costs 57.15\$ (38.10\$ for reward and 19.05\$ for AMT).
\section{Timing Details}
\label{appx:timing}
We report the time required for fine-tuning on the full amount of data for each subject/cloth-type combination in Tab.~\ref{tab:eval_timing}. Timings are measured on a single NVIDIA V100 GPU of our internal cluster. We fine-tune 256 epochs for each model. In comparison, NASA runs at a fixed number of iterations (200k) which usually takes more than 10 hours. SCANimate needs to train subject/cloth-type-specific skinning networks and neural SDFs which takes more than 10 hours and can be as much as 30 hours; the neural SDF training alone takes 1000 epochs per model. LEAP takes two days to train on all training and fine-tuning data. %, however its predictions are overly-smooth and not comparable to SCANimate and MetaAvatar.

\begin{table}
 \renewcommand{\tabcolsep}{3.0pt}
 \centering
 \begin{tabular}{ c c c c c }
 \hline
 \multicolumn{5}{c}{Per model fine-tuning time (hours)} \\ \hline
  00122-shortlong & 00122-shortshort & 00134-longlong & 00134-longshort & 00215-jerseyshort \\ \hline
  0.71 & 1.1 & 3.7 & 3.7 & 0.93 \\ \hline
  00215-longshort & 00215-poloshort & 00215-shortlong & 03375-blazerlong & 03375-longlong \\ \hline
  0.64 & 1.08 & 1.0 & 1.04 & 2.08 \\
 \hline
 \end{tabular}
 \vspace{0.05in}
 \caption{\textbf{Per-model fine-tuning time with 100\% data.} Our model runs with a fixed number of epochs, thus the fine-tuning time scales linearly with the amount of fine-tuning data. With 8 depth frames it only takes less than 2 minutes for fine-tuning.}
 \label{tab:eval_timing}
\end{table}

\section{Additional Qualitative Results}
\label{appx:additional_quatlitative}
In this section we show additional qualitative results, including 1) results on depth rendered from raw scans, 2) results on real depth images, 3) results on single raw scan, 4) comparison with pre-trained SCANimate model. All qualitative results are in video format and can be found in our project page: \href{https://neuralbodies.github.io/metavatar/}{\color{black}}{https://neuralbodies.github.io/metavatar/}

\subsection{Depth Rendered from Raw Scans}
\begin{figure}[t]
\centering
\includegraphics [width=\textwidth]{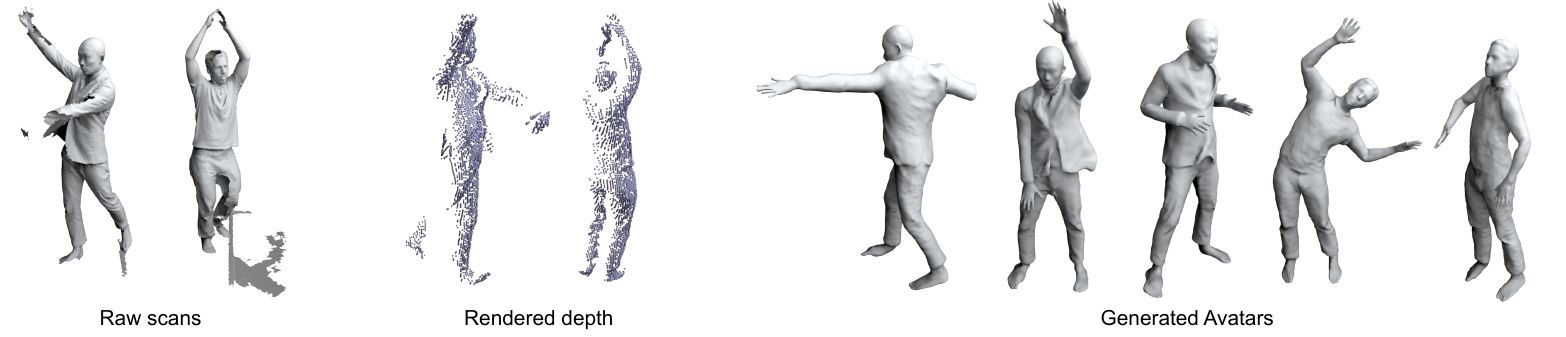}
\caption{Although our meta-model is learned on registered CAPE meshes, it generalizes well on depth images rendered from raw scans.}
\label{fig:qualitative_raw}
\end{figure}
To demonstrate that our meta-learned is robust to domain gaps of data, we render depth images from the raw scans of the CAPE~\cite{CAPE:CVPR:20} dataset and demonstrate that our meta-learned model, which was trained on registered meshes, generalizes well on raw scans. Note that raw scans often come with large holes, noisy surfaces, redundant background points, and missing parts (\eg\ left side of Fig.~\ref{fig:qualitative_raw}). To remove the background points such as floors and walls, we use the ground-truth CAPE meshes as a reference, removing any raw scan point whose distance is greater than 2cm from the corresponding ground truth clothed CAPE mesh. To handle the missing hands/feet problem, we replace them with hands and feet of the underlying SMPL registrations. We fine-tune our model on the 00122 subject with shortlong cloth-type and the 03375 subject with blazerlong cloth-type since we only have access to scans of these two subjects; for subject 00122 we fine-tuned on one action, while for 03375 we follow the same protocol which fine-tunes the model on trial1 sequences and test on trial2 sequences. We show several static frames in Fig.~\ref{fig:qualitative_raw}. 

\begin{figure}[t]
\centering
\includegraphics [width=\textwidth]{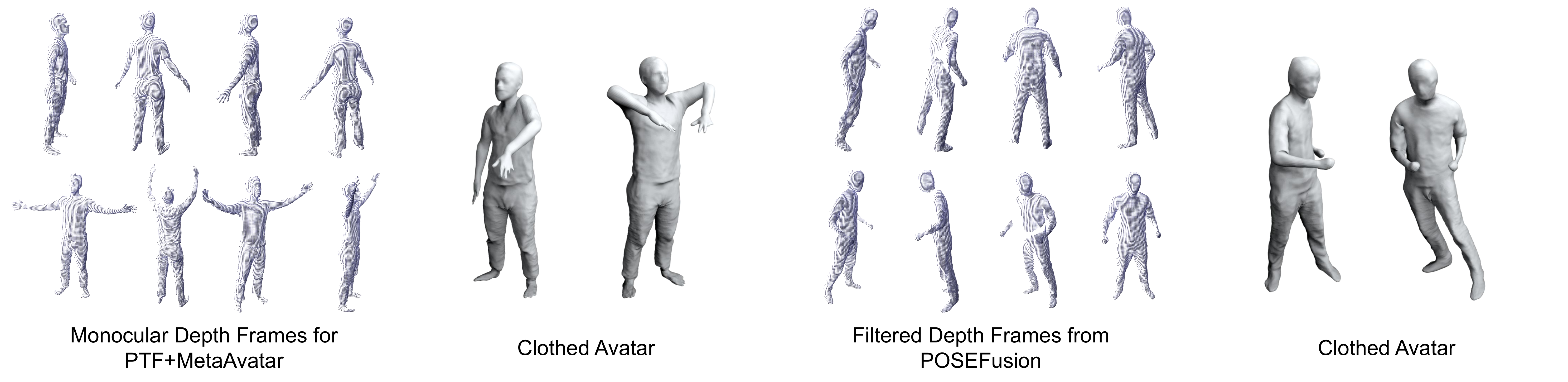}
\caption{\textbf{Left:} inputs to PTF and MetaAvatar and the generated avatar. \textbf{Right:} filtered and fused depth points from POSEFusion pipeline and the generated avatar.}
\label{fig:qualitative_ptf_posefusion}
\end{figure}

To further demonstrate the robustness of our approach to noisy SMPL registrations, we also utilize a recently released work PTF~\cite{PTF:CVPR:2021}, which is able to register SMPL to single-view point clouds, to register SMPL to 8 frames of single-view point clouds rendered from CAPE raw scans. Using these noisy registrations and rendered depth frames, we fine-tuned an avatar and show the pose extrapolation results in Fig.~\ref{fig:qualitative_ptf_posefusion}.

\subsection{Real Depth Images}
Real depth sensors often give noisy outputs, and it is necessary to use tracking and fusion techniques to filter out noise and outliers. We thus utilize POSEFusion~\cite{li2021posefusion} to obtain the necessary data for creating avatars. The input to POSEFusion is a monocular RGBD video of a clothed person moving and rotating, showing both his/her frontal and back views. It uses tracking and fusion, guided by SMPL estimations, to fuse invisible parts from future frames to current frames, such that it can reconstruct the full-body mesh at each RGBD frame. Given the reconstructed full-body meshes as well as SMPL registrations, we render the first 8 frames of reconstructions and use these 8 frames along with their estimated SMPL fits to create our avatar. The avatar is then animated with estimated SMPL registrations of the rest of the sequence (~210 frames) and sample poses from CAPE dataset (~140 frames). Sample results are illustrated in Fig.~\ref{fig:qualitative_ptf_posefusion}.

\subsection{Single Scan Animation}
\begin{figure}[t]
\centering
\includegraphics [width=\textwidth]{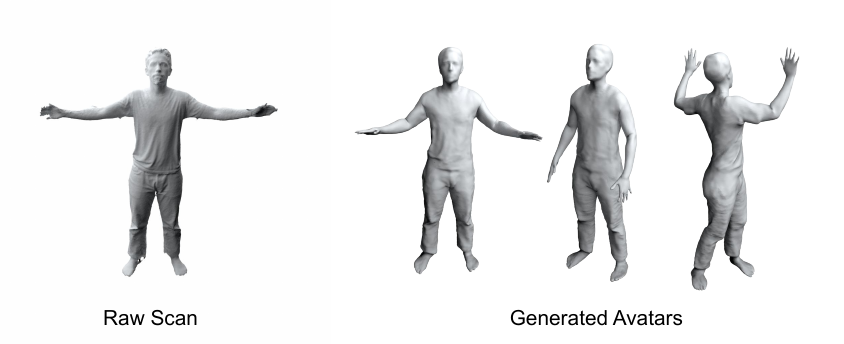}
\caption{\textbf{Single Scan Animation}. Our meta-learned model is able to animate a single scan.}
\label{fig:qualitative_single_scan}
\end{figure}
To additionally demonstrate the robustness of our meta-learned model, we also fine-tuned an avatar using a single full-body scan. Sample results are showed in Fig.~\ref{fig:qualitative_single_scan}. Note that our model is meta-learned on rendered depth images yet it also yields reasonable results on full-body scans.

\subsection{Comparison with Pre-trained SCANimate Model}
\begin{figure}[t]
\centering
  \begin{subfigure}{0.16\textwidth}
  \captionsetup{labelformat=empty, font=scriptsize}
    \includegraphics [trim=12cm 6cm 13cm 5cm, width=\textwidth]{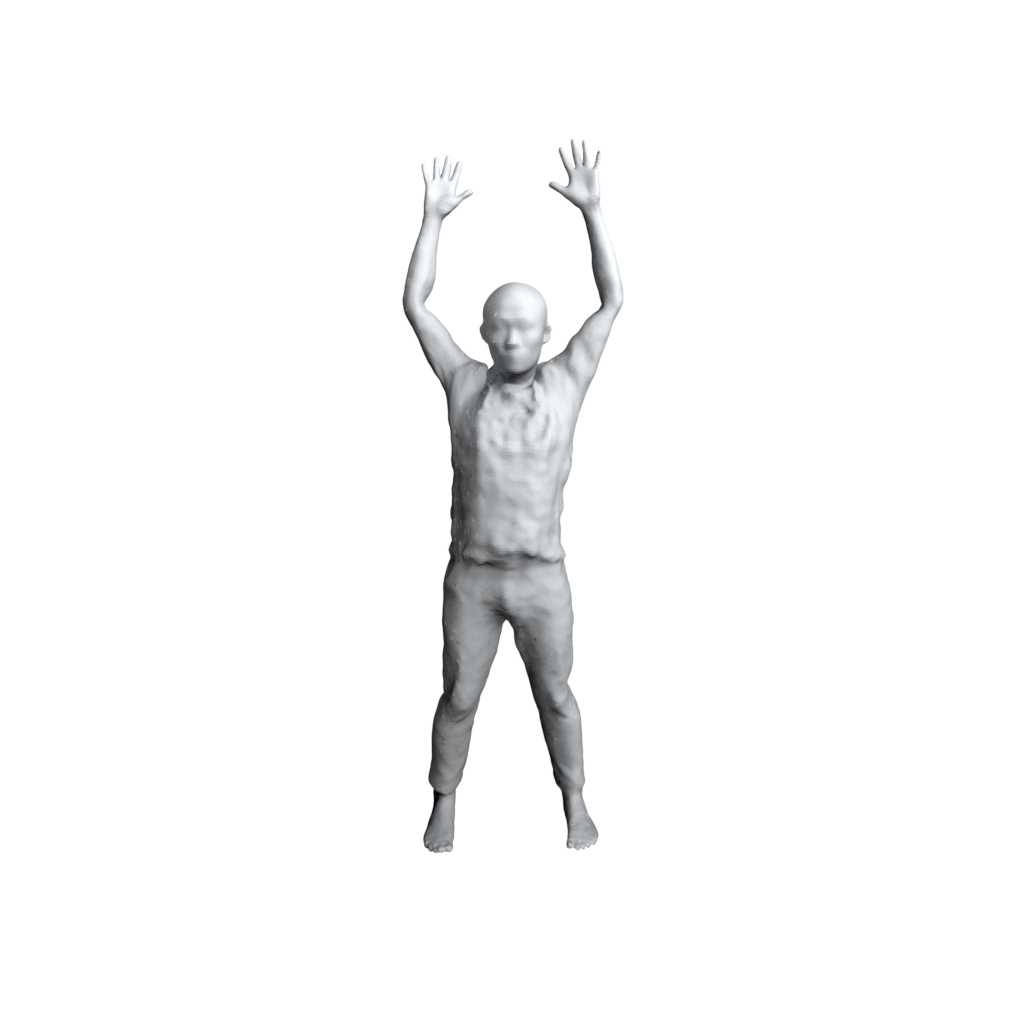}
  \caption{SCANimate, 16 scans}
  \end{subfigure}
  \begin{subfigure}{0.16\textwidth}
  \captionsetup{labelformat=empty, font=scriptsize}
    \includegraphics [trim=12cm 6cm 13cm 5cm, width=\textwidth]{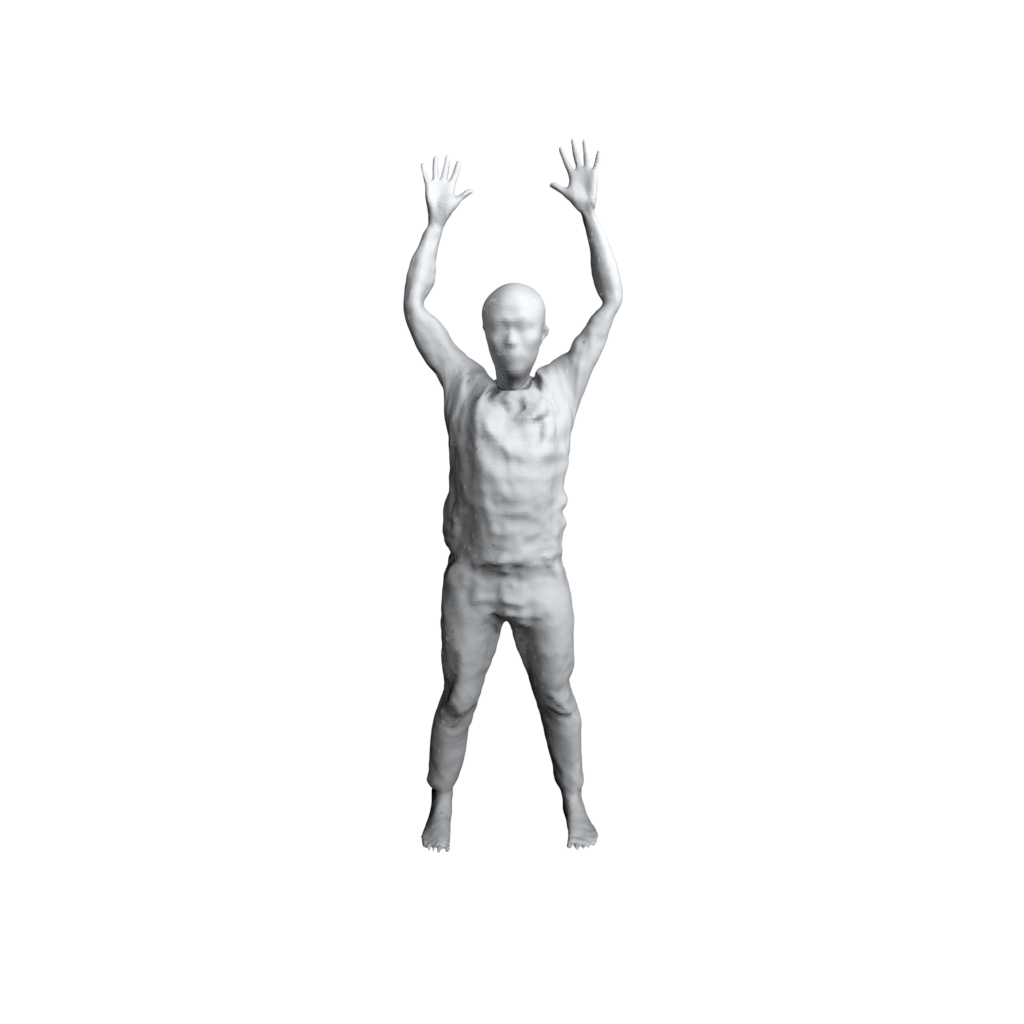}
  \caption{SCANimate, 8 scans}
  \end{subfigure}
  \begin{subfigure}{0.16\textwidth}
  \captionsetup{labelformat=empty, font=scriptsize}
    \includegraphics [trim=12cm 6cm 13cm 5cm, width=\textwidth]{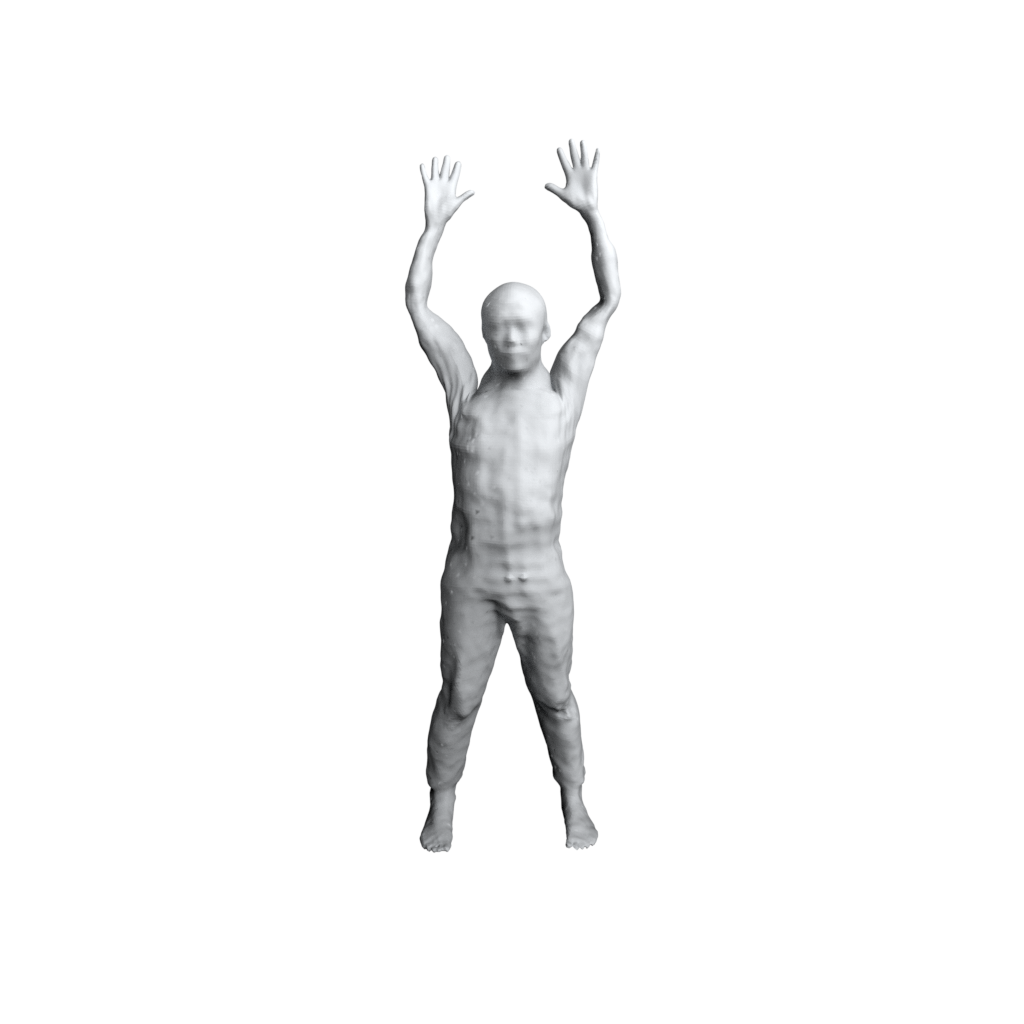}
  \caption{SCANimate, 1 scan}
  \end{subfigure}
  \begin{subfigure}{0.16\textwidth}
  \captionsetup{labelformat=empty, font=scriptsize}
    \includegraphics [trim=12cm 6cm 13cm 5cm, width=\textwidth]{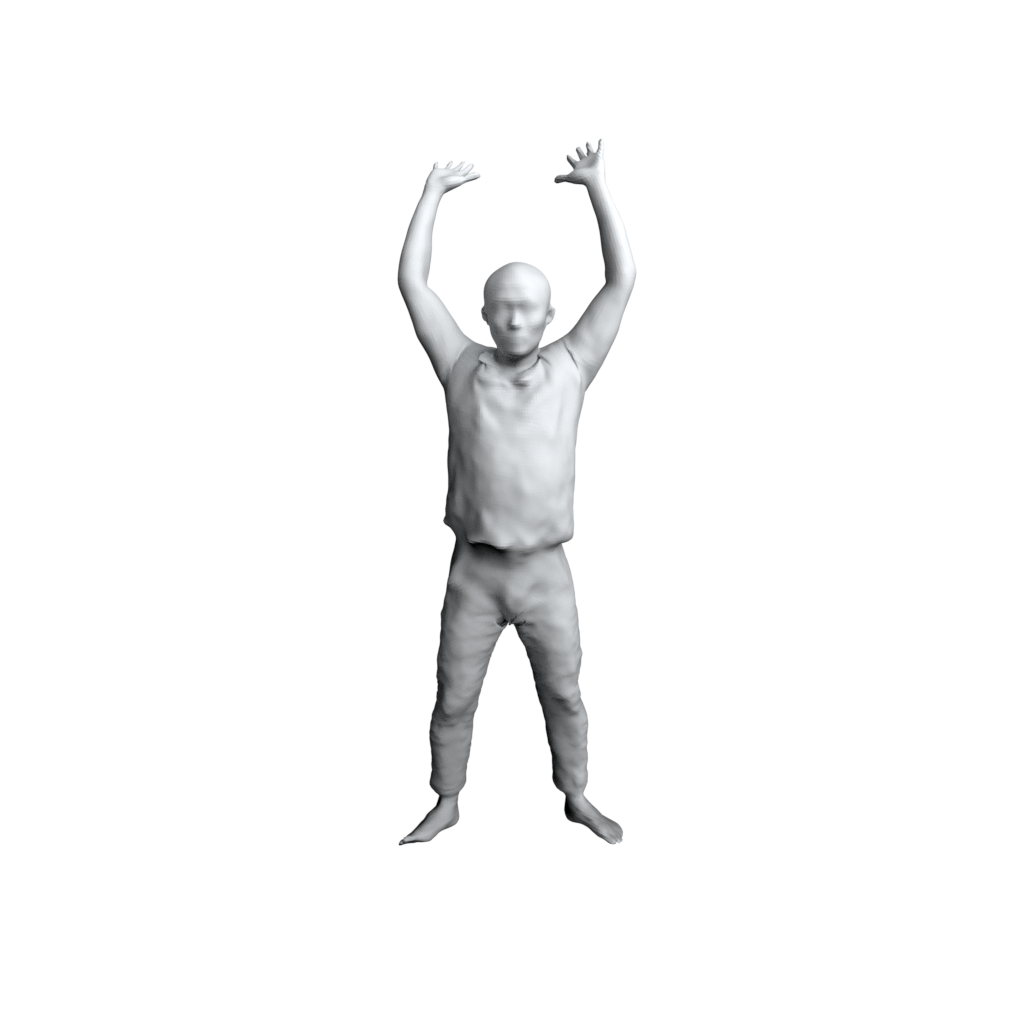}
  \caption{Ours, 16 scans}
  \end{subfigure}
  \begin{subfigure}{0.16\textwidth}
  \captionsetup{labelformat=empty, font=scriptsize}
    \includegraphics [trim=12cm 6cm 13cm 5cm, width=\textwidth]{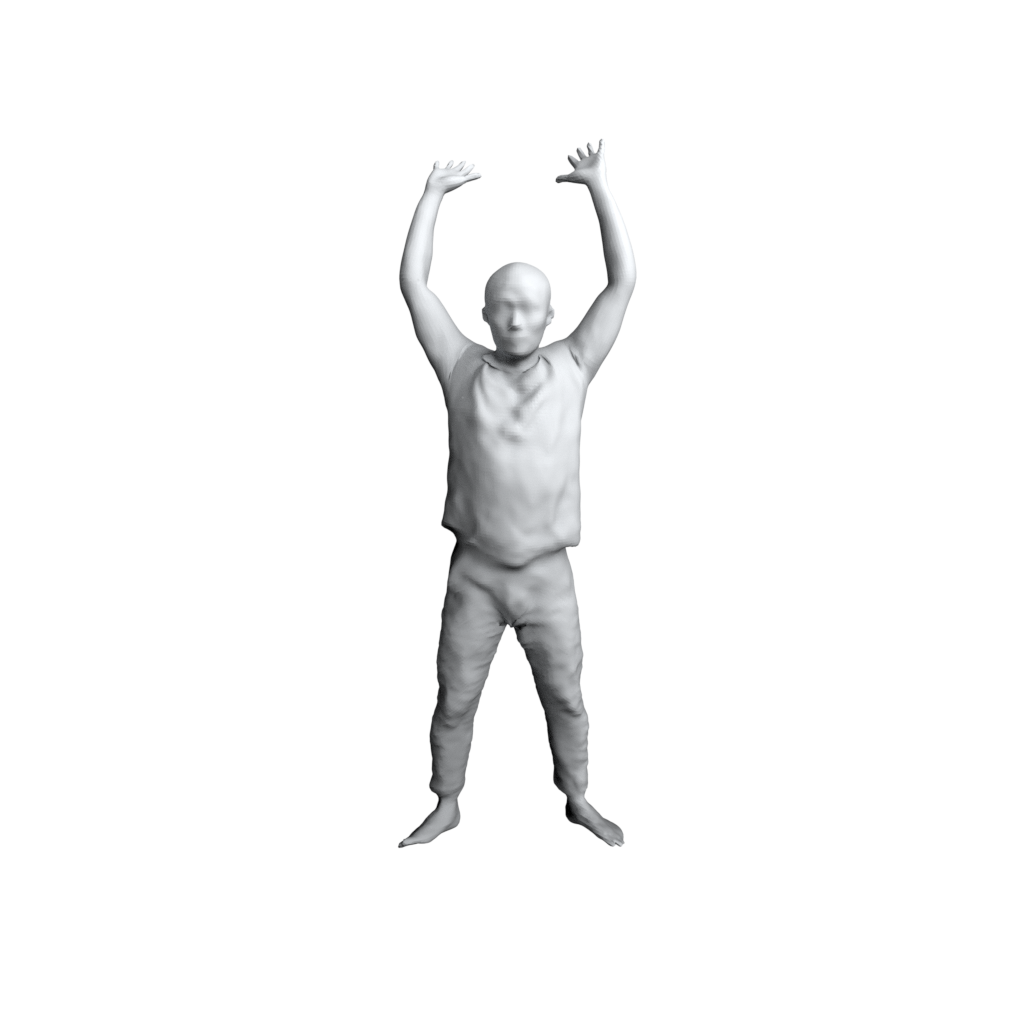}
  \caption{Ours, 8 scans}
  \end{subfigure}
  \begin{subfigure}{0.16\textwidth}
  \captionsetup{labelformat=empty, font=scriptsize}
    \includegraphics [trim=12cm 6cm 13cm 5cm, width=\textwidth]{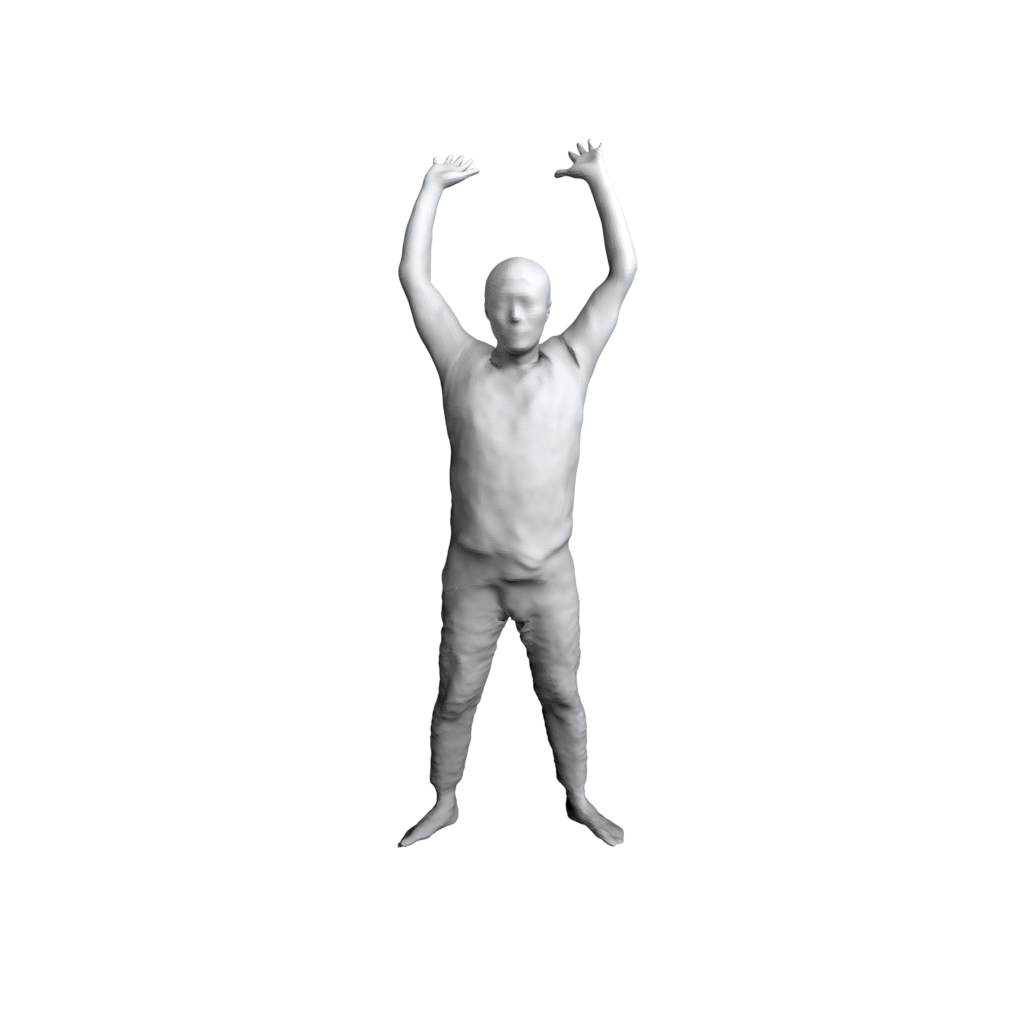}
  \caption{Ours, 1 scan}
  \end{subfigure}
\caption{\textbf{Comparison with SCANimate on limited data}. Note that SCANimate sets hands and feet poses to 0. With limited data, SCANimate cannot model cloth deformation properly and has obvious artifacts around elbows. In comparison, our model can produce reasonable cloth deformations and body shapes with limited data.}
\label{fig:qualitative_scanimate_comp}
\end{figure}
To demonstrate the advantage of meta-learning, we also compare our results with results fine-tuned from pre-trained SCANimate model. We use the official SCANimate release code, which comes with 16 training raw scans of subject 03375-shortlong and several pre-trained models on different subject/cloth-type combinations. We found that 00096-shirtlong is in our training set and the model has similar body shape and cloth-type to 03375-shortlong. We thus fine-tune the pre-trained model of 00096-shirtlong with provided raw scans of subject 03375-shortlong, using the default configuration of SCANimate.

We fine-tune SCANimate with 16/8/1 raw scan(s) to verify its performance on reduced data and compare it to our model. We animate the fine-tuned avatars with 03375-shortlong’s trial2 action sequences; these actions have not been seen in either training data or fine-tuning data. Sample results are shown in Fig.~\ref{fig:qualitative_scanimate_comp}.
\section{Limitations}
\label{appx:limitations}
\begin{figure}[t]
\centering
  \begin{subfigure}{0.16\textwidth}
  \captionsetup{labelformat=empty, font=scriptsize}
    \includegraphics [trim=10cm 8cm 11cm 5cm, width=\textwidth]{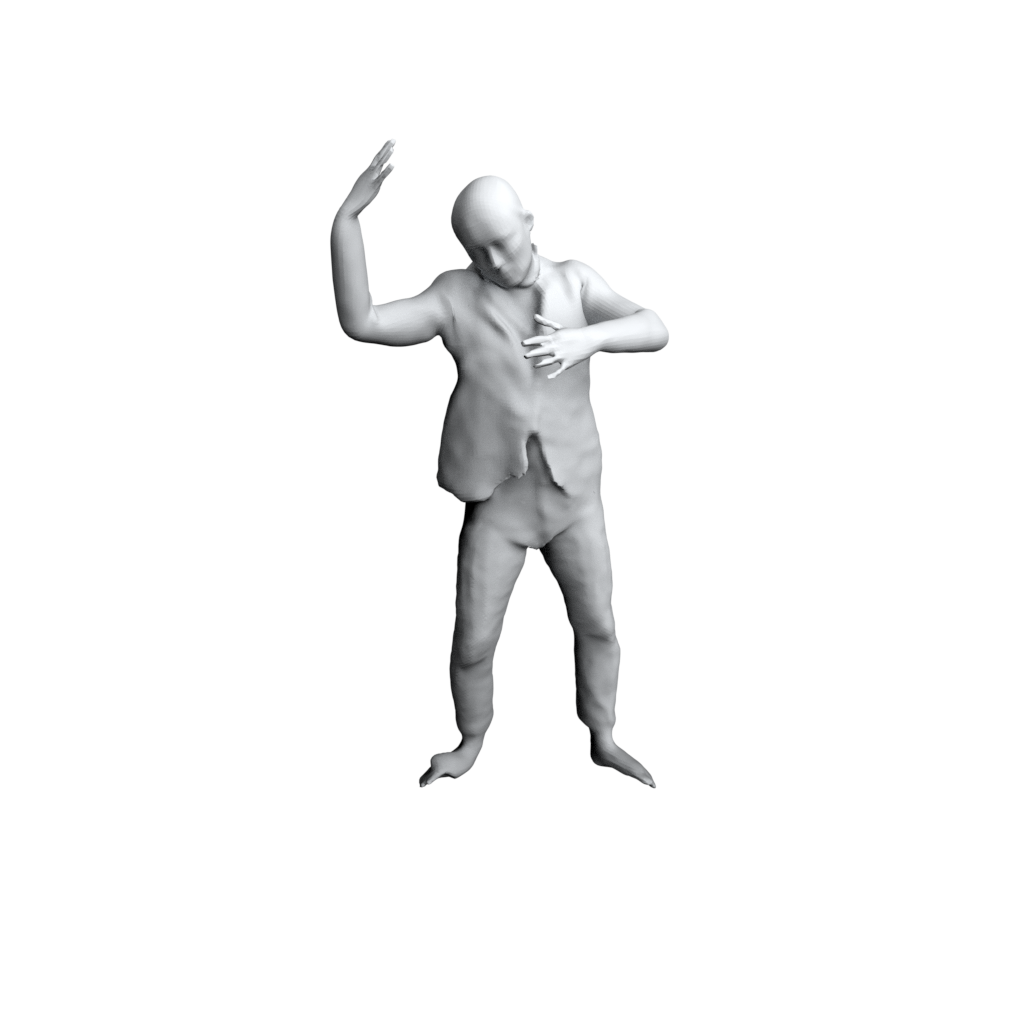}
  \end{subfigure}
  \begin{subfigure}{0.16\textwidth}
  \captionsetup{labelformat=empty, font=scriptsize}
    \includegraphics [trim=10cm 8cm 11cm 5cm, width=\textwidth]{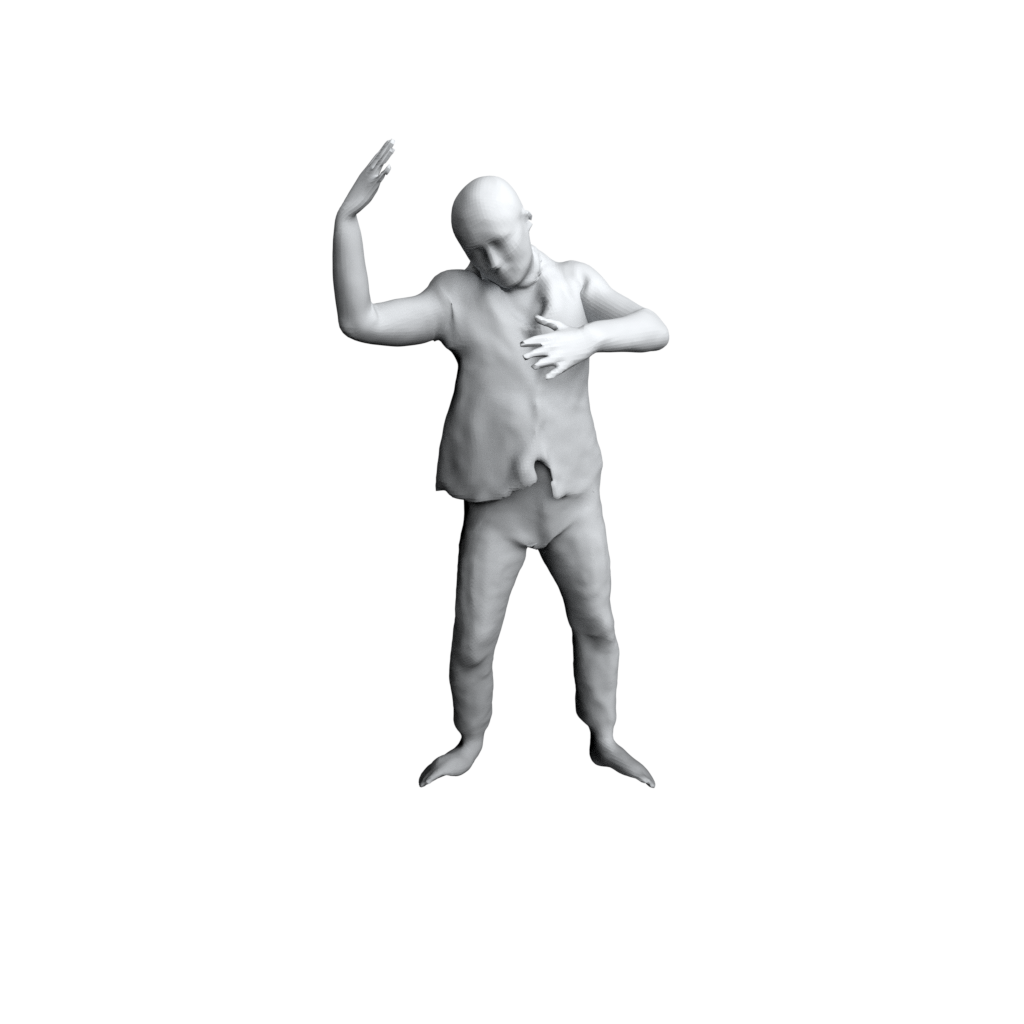}
  \end{subfigure}
  \begin{subfigure}{0.16\textwidth}
  \captionsetup{labelformat=empty, font=scriptsize}
    \includegraphics [trim=10cm 8cm 11cm 5cm, width=\textwidth]{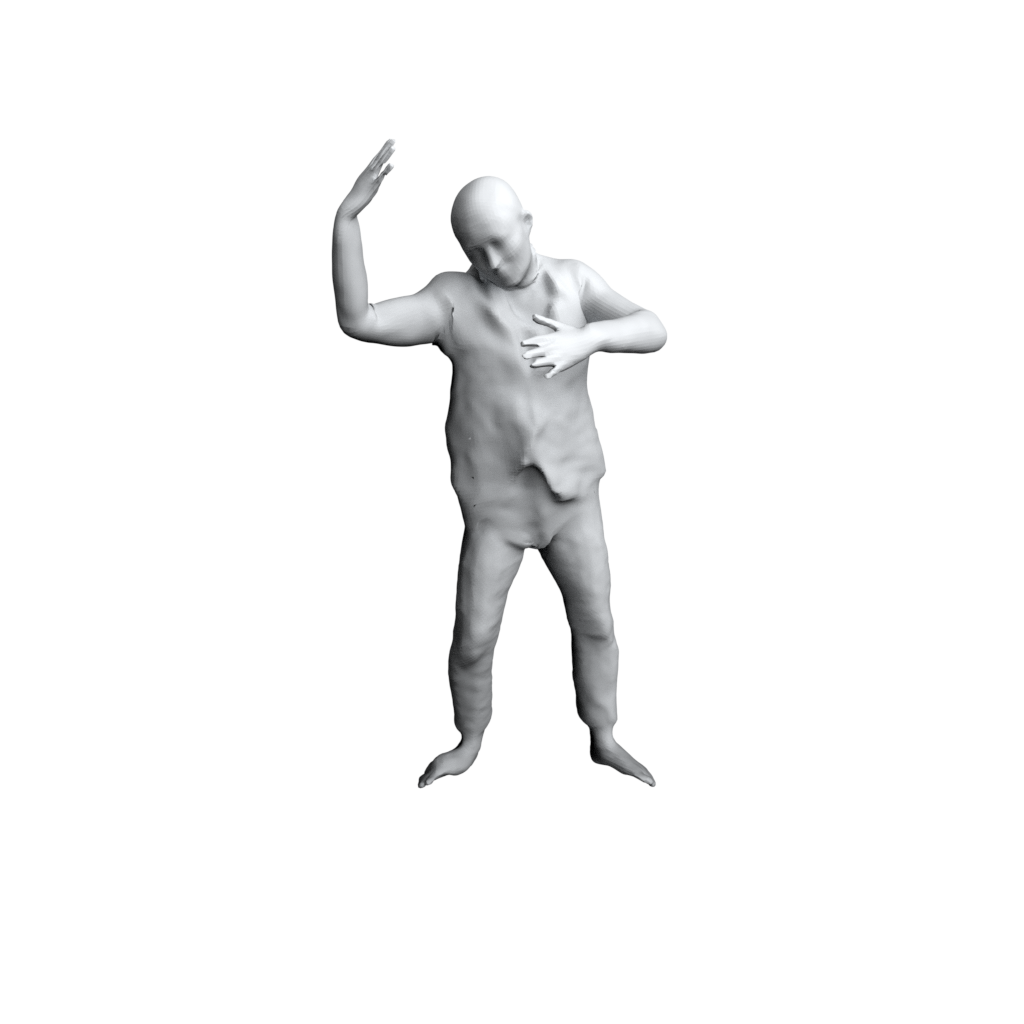}
  \end{subfigure}
  \begin{subfigure}{0.16\textwidth}
  \captionsetup{labelformat=empty, font=scriptsize}
    \includegraphics [trim=10cm 8cm 11cm 5cm, width=\textwidth]{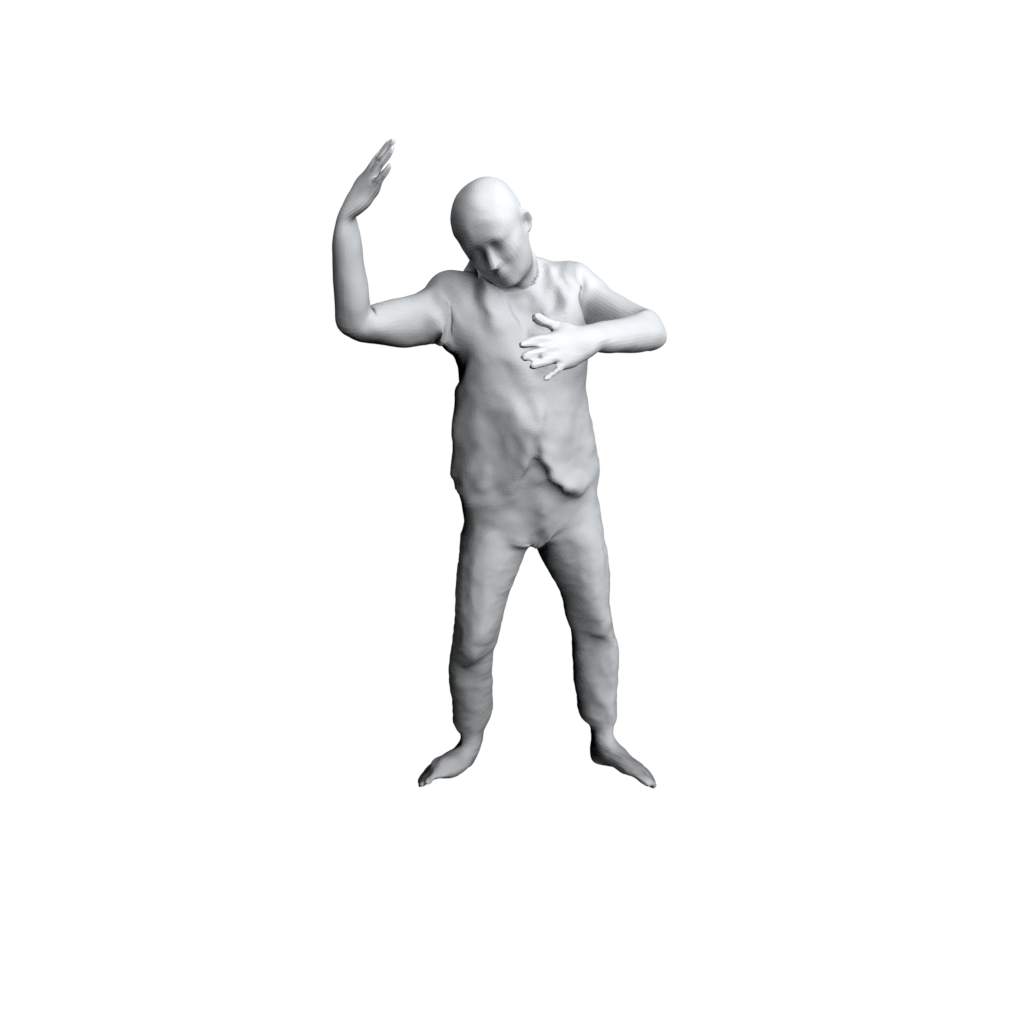}
  \end{subfigure}
  \begin{subfigure}{0.16\textwidth}
  \captionsetup{labelformat=empty, font=scriptsize}
    \includegraphics [trim=10cm 8cm 11cm 5cm, width=\textwidth]{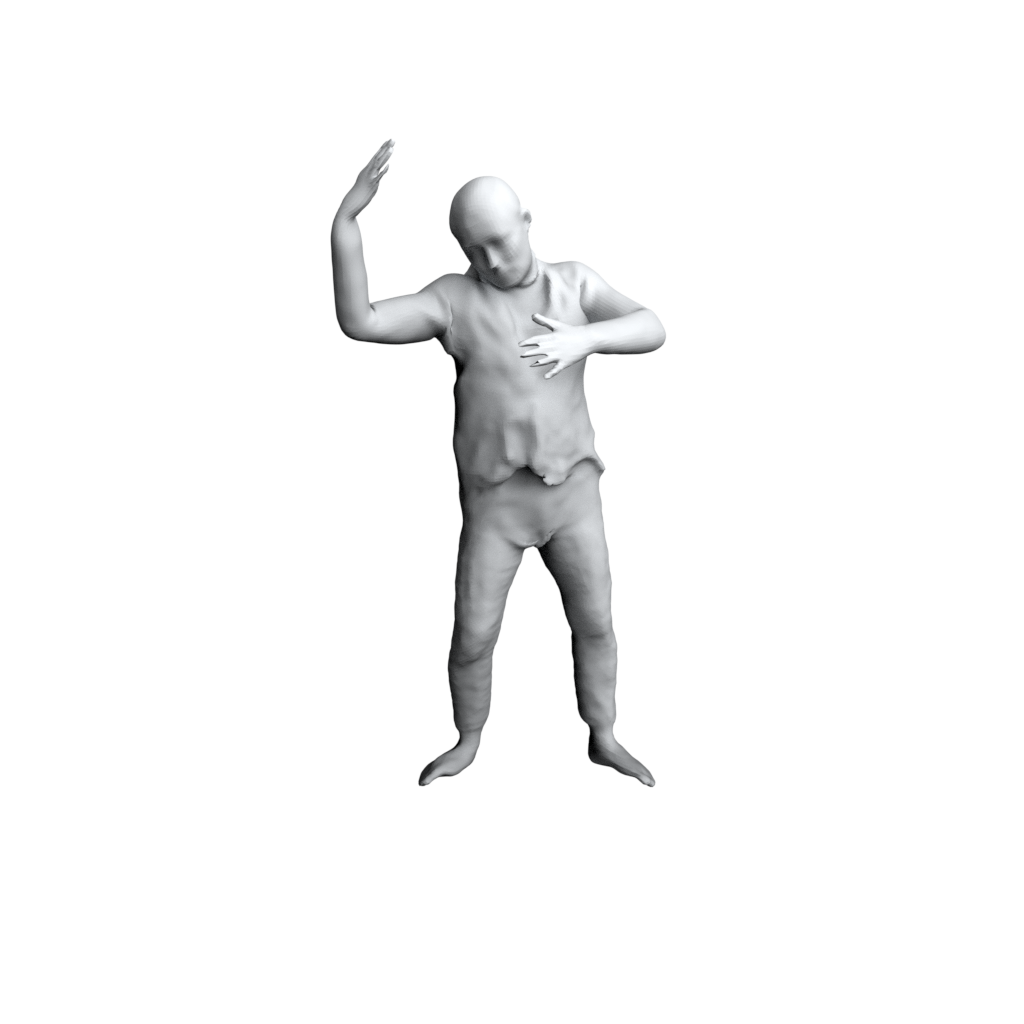}
  \end{subfigure}
  \begin{subfigure}{0.16\textwidth}
  \captionsetup{labelformat=empty, font=scriptsize}
    \includegraphics [trim=10cm 8cm 11cm 5cm, width=\textwidth]{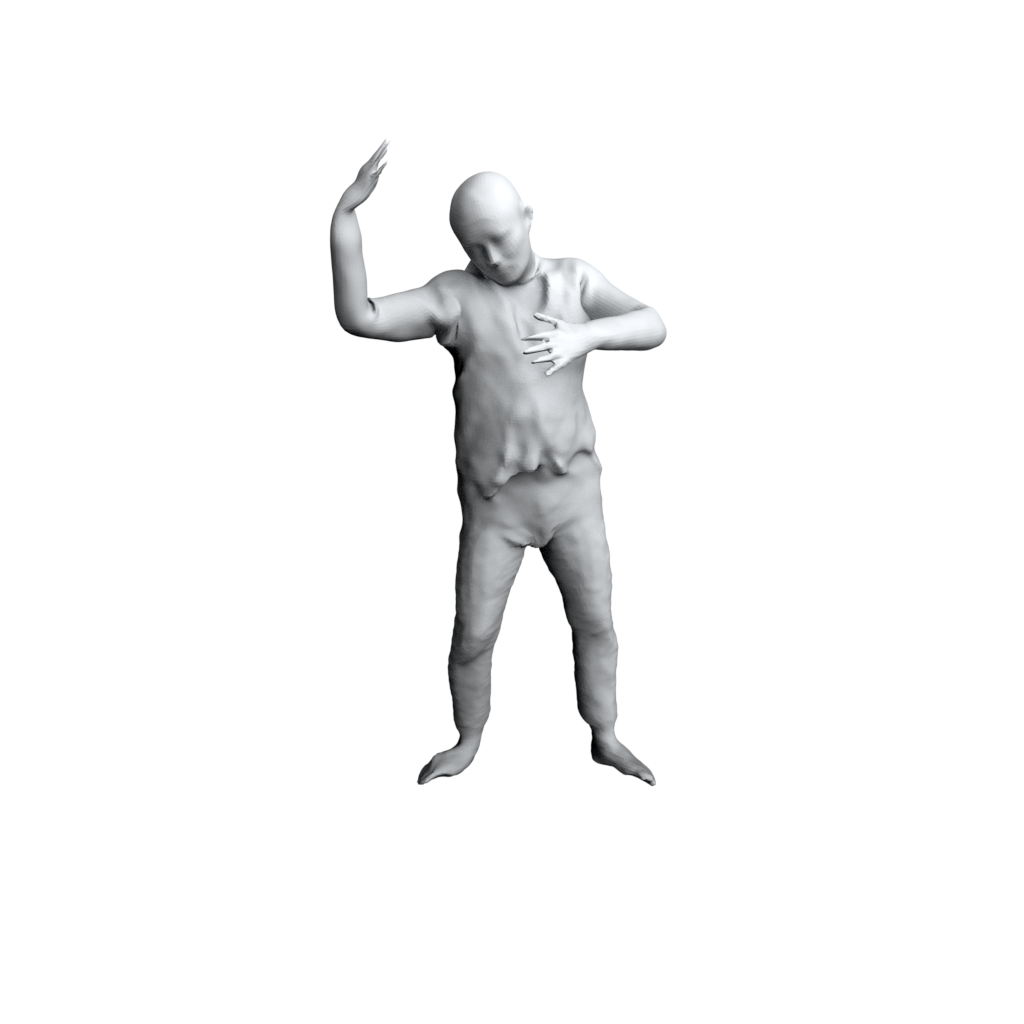}
  \end{subfigure}
  \begin{subfigure}{0.16\textwidth}
  \captionsetup{labelformat=empty, font=scriptsize}
    \includegraphics [trim=12cm 8cm 12cm 5cm, width=\textwidth]{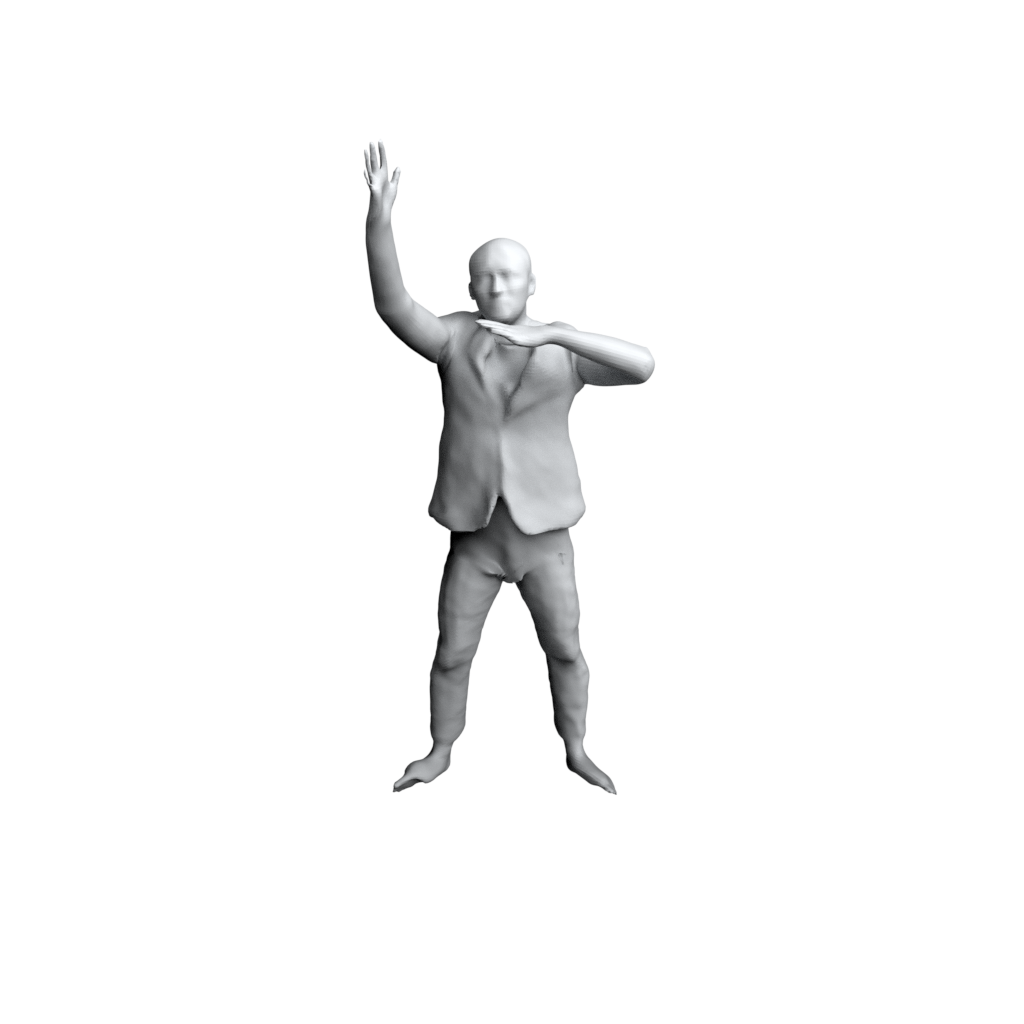}
  \caption{100\%}
  \end{subfigure}
  \begin{subfigure}{0.16\textwidth}
  \captionsetup{labelformat=empty, font=scriptsize}
    \includegraphics [trim=12cm 8cm 12cm 5cm, width=\textwidth]{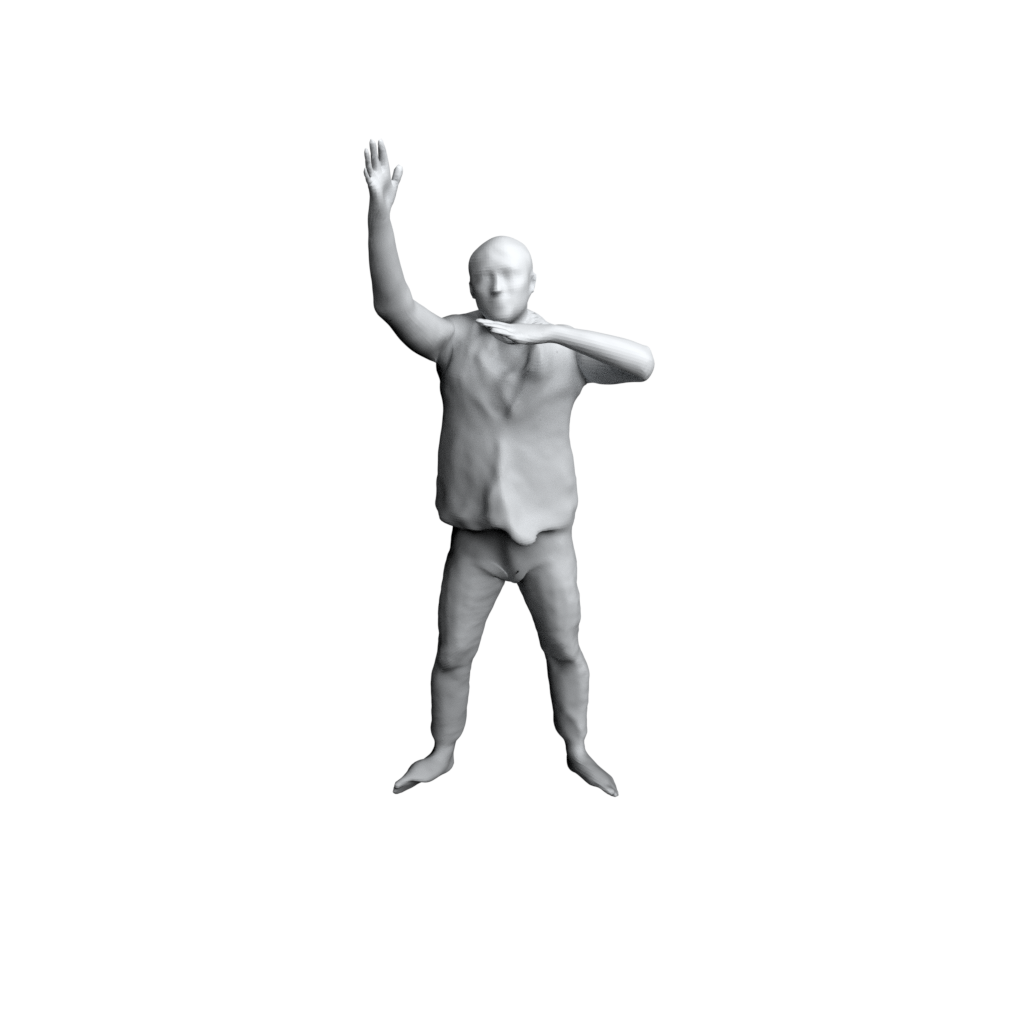}
  \caption{50\%}
  \end{subfigure}
  \begin{subfigure}{0.16\textwidth}
  \captionsetup{labelformat=empty, font=scriptsize}
    \includegraphics [trim=12cm 8cm 12cm 5cm, width=\textwidth]{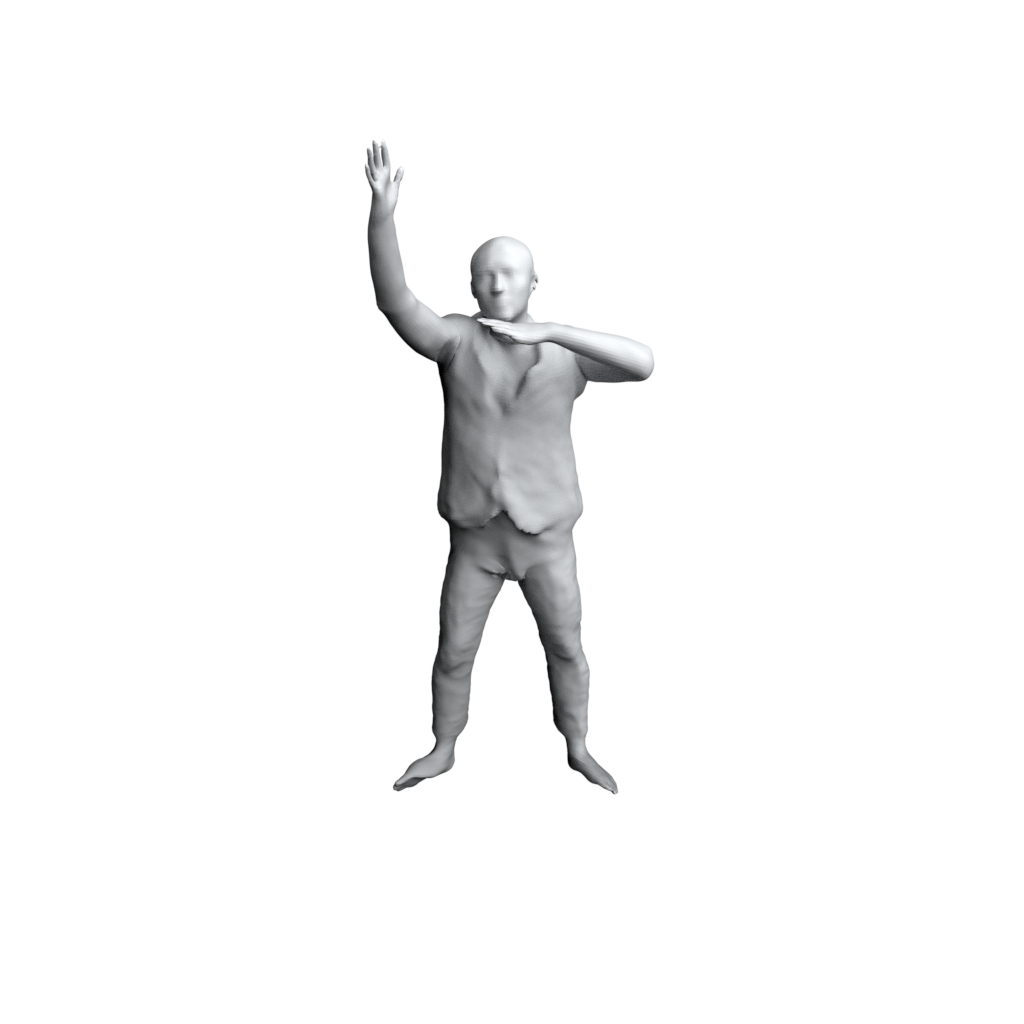}
  \caption{20\%}
  \end{subfigure}
  \begin{subfigure}{0.16\textwidth}
  \captionsetup{labelformat=empty, font=scriptsize}
    \includegraphics [trim=12cm 8cm 12cm 5cm, width=\textwidth]{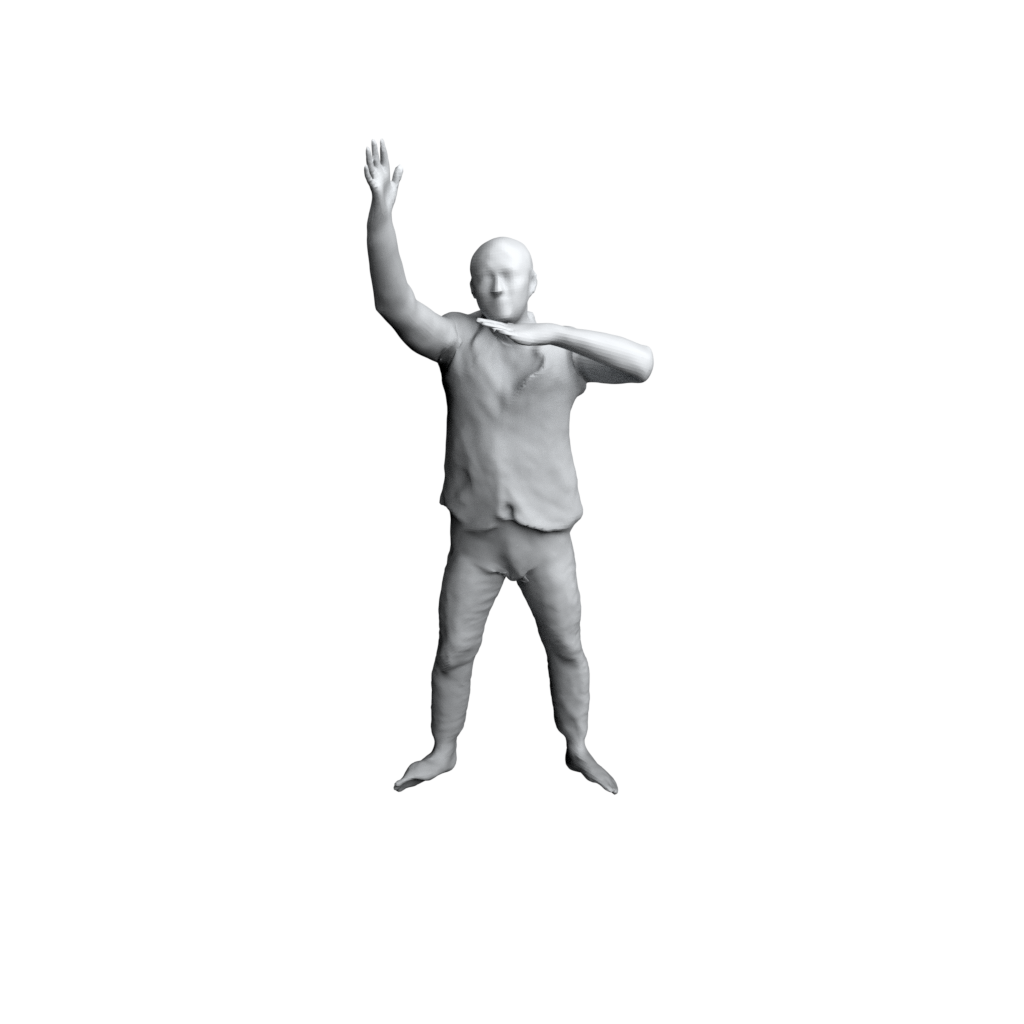}
  \caption{10\%}
  \end{subfigure}
  \begin{subfigure}{0.16\textwidth}
  \captionsetup{labelformat=empty, font=scriptsize}
    \includegraphics [trim=12cm 8cm 12cm 5cm, width=\textwidth]{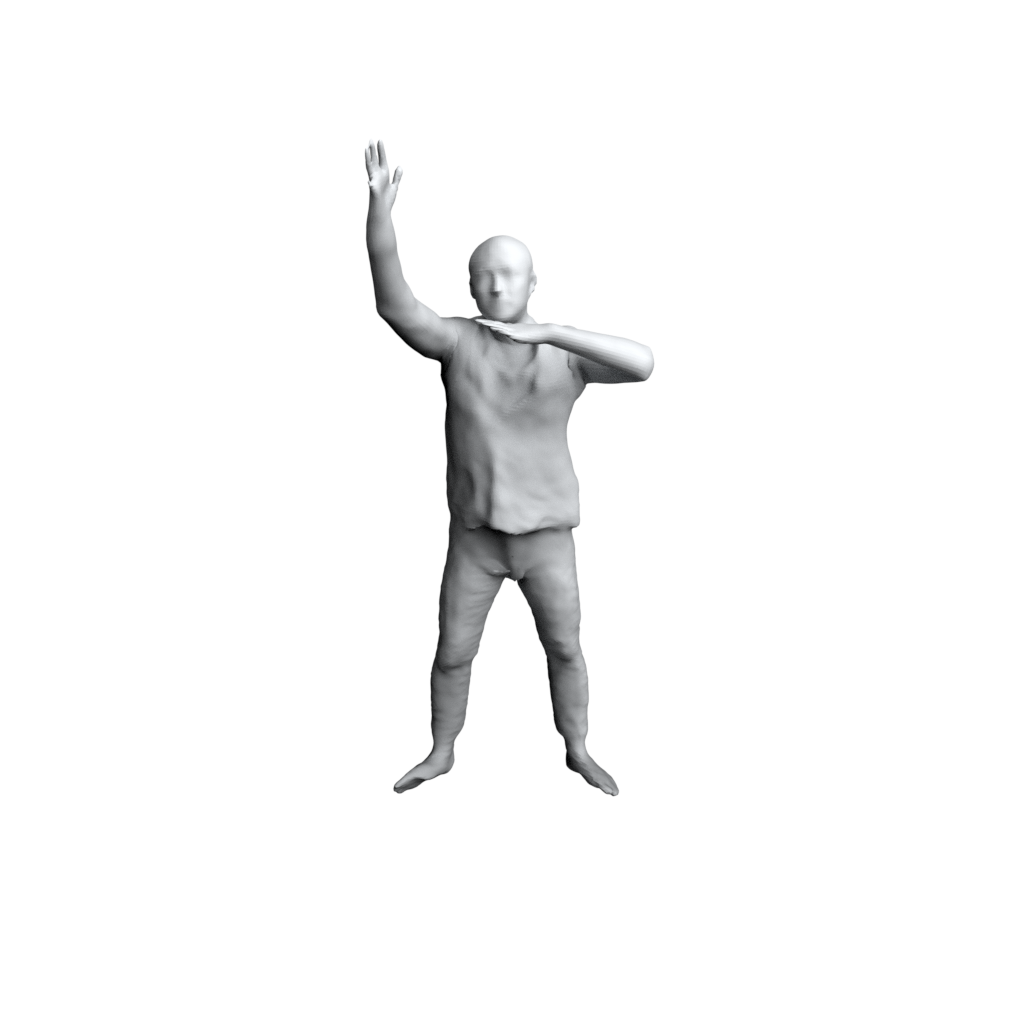}
  \caption{5\%}
  \end{subfigure}
  \begin{subfigure}{0.16\textwidth}
  \captionsetup{labelformat=empty, font=scriptsize}
    \includegraphics [trim=12cm 8cm 12cm 5cm, width=\textwidth]{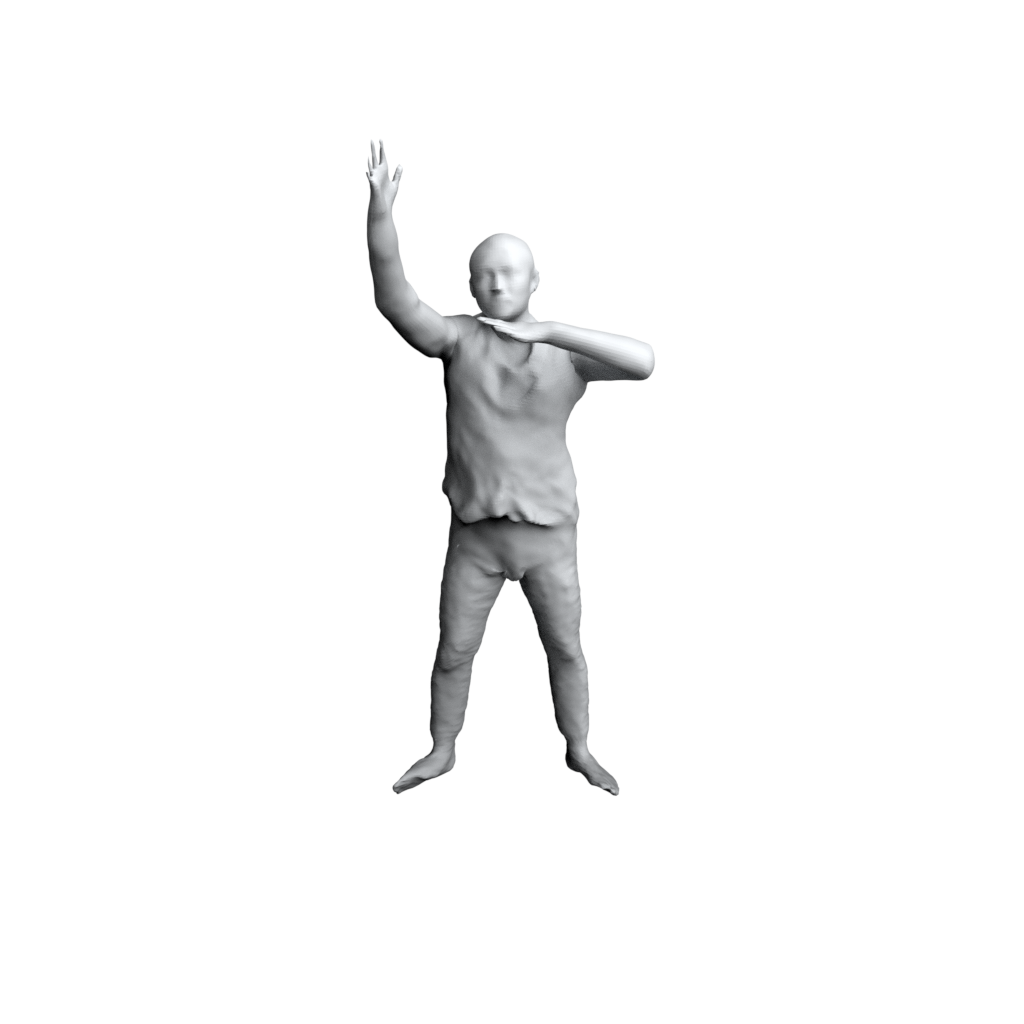}
  \caption{<1\%}
  \end{subfigure}
\caption{\textbf{Limitation}. For the challenging blazer outfit which is not seen during meta-learning, our model fails to capture its dynamics with limited fine-tuning data.}
\label{fig:failure_cases}
\end{figure}

As illustrated in Fig.~\ref{fig:failure_cases},
the proposed MetaAvatar can generalize to the blazer outfit which is unseen during meta-learning. 
However, the model needs to see a moderate amount of data in order to represent the cloth dynamics of this outfit, otherwise it cannot capture the specific dynamics such as the tails of the blazer.

Another limitation is that MetaAvatar (and all other baselines) relies on accurate SMPL registrations to input point clouds. In the future, we aim to combine our work with the state-of-the-art parametric human body registration approaches~\cite{PTF:CVPR:2021,bhatnagar2020loopreg} to enable the joint learning and optimization of neural avatars and SMPL registrations. 
\end{document}